%% file: main.tex
\documentclass[10pt]{article}

\input{./includes/macro}

\input{./includes/macro-identifiable}

\title{Artificial Intelligence to Assess Dental Findings from Panoramic Radiographs -- A Multinational Study}

\input{./includes/preamble-author}
\date{}

\begin{document}
\maketitle

\input{./includes/section-abstract}
\newpage

\input{./includes/section-introduction}
\input{./includes/section-materials-and-methods}
\input{./includes/section-results}
\input{./includes/section-discussion}

\newpage

\section*{Supplementary Material}
\setcounter{section}{0}
\renewcommand{\thesection}{\Alph{section}}
\input{./includes/section-supplementary-material}
\newpage

\bibliography{main}

\end{document}

%% file: includes/macro.tex
\usepackage{authblk}

\usepackage[a4paper, margin=1in]{geometry}
\usepackage[pagebackref, colorlinks]{hyperref}
\usepackage[round, comma, numbers, sort&compress]{natbib}
\usepackage[dvipsnames]{xcolor}

\usepackage[inline]{enumitem}
\usepackage[T1]{fontenc}
\usepackage{libertine}  
\usepackage[libertine]{newtxmath}
\usepackage{relsize}
\usepackage{siunitx}
\usepackage{xspace}
\usepackage{booktabs}
\usepackage{caption}
\usepackage{colortbl}

\usepackage{graphicx}

\def\eg{\emph{e.g.}}
\def\ie{\emph{i.e.}}

\definecolor{table-bg}{RGB}{214, 229, 226}

\def\objness{s}
\def\objProb{\mathbf{q}}
\def\objMap{\mathbf{M}^\mathrm{obj}}

\def\semProb{\mathbf{p}}
\def\semMap{\mathbf{M}^\mathrm{sem}}

\def\func{\pi}
\def\corr{r}
\def\zeroOne{\param{0, 1}}

\newcommand{\para}[1]{{\left(#1\right)}}
\newcommand{\param}[1]{{\left[#1\right]}}
\newcommand{\paral}[1]{{\left\{#1\right\}}}
\newcommand{\paraa}[1]{{\left\langle#1\right\rangle}}


%% file: includes/macro-identifiable.tex
\newcommand{\hospitalN}{\textit{Radboud University Medical Center}\xspace}
\newcommand{\hospitalB}{\textit{Proradis} (a Brazilian company)\xspace}
\newcommand{\hospitalT}{\textit{National Taiwan University Hospital}\xspace}

\def\gdpYW{Y.W.\xspace}
\def\gdpCC{C.C.\xspace}
\def\gdpHC{H.C.\xspace}
\def\gdpHW{H.W.\xspace}

%% file: includes/preamble-author.tex
\author[1]{Yin-Chih Chelsea Wang, DDS}
\author[7]{Tsao-Lun Chen, MSc}
\author[6]{Shankeeth Vinayahalingam, MD, DMD}
\author[10]{Tai-Hsien Wu, PhD}
\author[3,4]{Chu Wei Chang, DDS}
\author[3,4]{Hsuan Hao Chang, DDS}
\author[3,4]{Hung-Jen Wei, DDS}
\author[4]{Mu-Hsiung Chen, MS}
\author[10]{Ching-Chang Ko, DDS, PhD}
\author[5]{David Anssari Moin, PhD}
\author[6]{Bram van Ginneken, PhD, MSc}
\author[6,8]{Tong Xi, PhD, DMD, MD}
\author[3,4]{Hsiao-Cheng Tsai, DDS, PhD}
\author[3,4]{Min-Huey Chen, DDS, PhD}
\author[2,9]{Tzu-Ming Harry Hsu, PhD}
\author[1,4,*]{HYE Chou, PhD}

\affil[1]{Graduate Institute of Oral Biology, College of Medicine, National Taiwan University. No. 1, Changde St., Taipei City 100, Taiwan}
\affil[2]{Massachusetts Institute of Technology. 77 Massachusetts Avenue, Cambridge, MA 02139, USA}
\affil[3]{Graduate Institute of Clinical Dentistry, College of Medicine, National Taiwan University. No. 1, Changde St., Taipei City 100, Taiwan}
\affil[4]{Department of Dentistry, National Taiwan University Hospital. No. 1, Changde St., Taipei City 100, Taiwan}
\affil[5]{Promaton. Kon. Wilhelminaplein 1, 1062 HG Amsterdam, Netherlands}
\affil[6]{Radboud University Medical Center. Geert Grooteplein Zuid 10, 6525 GA Nijmegen, Netherlands}
\affil[7]{National Taiwan University of Science and Technology. No. 43, Section 4, Keelung Rd, Taipei City 106, Taiwan}
\affil[8]{Department of Maxillofacial Research, Clinical Institute, Faculty of Health, University of Southern, Odense, Denmark.}
\affil[9]{International Academia of Biomedical Innovation Technology. 3F., No. 764, Sec. 5, Zhongxiao E. Rd., Xinyi Dist., Taipei City 110039, Taiwan}
\affil[10]{The Ohio State University. 305 W 12th Ave, Columbus, OH 43210, USA}
\affil[*]{Corresponding Author: \href{mailto:hyechou@ntu.edu.tw}{\url{hyechou@ntu.edu.tw}}}

%% file: includes/section-abstract.tex
\begin{abstract}
\mbox{}\par\vspace{-\baselineskip}%


\paragraph{Background}
Dental panoramic radiographs (DPRs) are widely used in clinical practice for comprehensive oral assessment but present challenges due to overlapping structures and time constraints in interpretation.

\paragraph{Purpose}
This study aimed to establish a solid baseline for the AI-automated assessment of findings in DPRs by developing, evaluating an AI system, and comparing its performance with that of human readers across multinational data sets.

\paragraph{Materials and Methods}
We analyzed $\num{6669}$ DPRs from three data sets (the Netherlands, Brazil, and Taiwan), focusing on 8 types of dental findings.
The AI system combined object detection and semantic segmentation techniques for per-tooth finding identification.
Performance metrics included sensitivity, specificity, and area under the receiver operating characteristic curve (AUC-ROC).
AI generalizability was tested across data sets, and performance was compared with human dental practitioners.

\paragraph{Results}
The AI system demonstrated comparable or superior performance to human readers, particularly +67.9\% (95\% CI: 54.0\%--81.9\%; $p < .001$) sensitivity for identifying periapical radiolucencies and +4.7\% (95\% CI: 1.4\%--8.0\%; $p = .008$) sensitivity for identifying missing teeth.
The AI achieved a macro-averaged AUC-ROC of 96.2\% (95\% CI: 94.6\%--97.8\%) across 8 findings.
AI agreements with the reference were comparable to inter-human agreements in 7 of 8 findings except for caries ($p=.024$).
The AI system demonstrated robust generalization across diverse imaging and demographic settings and processed images 79 times faster (95\% CI: 75--82) than human readers.

\paragraph{Conclusion}
The AI system effectively assessed findings in DPRs, achieving performance on par with or better than human experts while significantly reducing interpretation time.
These results highlight the potential for integrating AI into clinical workflows to improve diagnostic efficiency and accuracy, and patient management.
Future work should focus on validating AI systems in real-world deployments.

\end{abstract}

%% file: includes/section-introduction.tex
\section{Introduction}

Dental panoramic radiographs (DPRs), also known as orthopantomograms (OPG), provide an overview of the oral region and are widely used as a primary diagnostic tool by dentists and oral and maxillofacial (OMF) surgeons in routine clinical practice.
DPRs allow for the detection of diagnostic findings, such as dental caries and lesions, as well as the assessment of prior treatments, including root canal fillings and dental implants.
These radiographs are particularly valued due to their low radiation exposure, brief acquisition times, and cost-effectiveness, enabling early detection of oral diseases for improved prognostic outcomes~\citep{choi2011assessment} and enhanced assessment of treatment statuses.
However, accurately interpreting DPRs is challenging due to the overlapping structures of tissues and bones~\citep{fourcade2019deep}. 
Furthermore, clinicians often perform interpretations under significant time constraints, focusing predominantly on symptomatic areas.
Studies have consequently reported low inter-observer agreement even among experienced dentists~\citep{kweon2018panoramic}, which can lead to suboptimal diagnostic performance in clinical scenarios~\citep{plessas2019impact}.

With the advent of artificial intelligence (AI), significant advancements have been made in developing computer-aided detection (CAD) systems to interpret panoramic radiographs for a variety of dental findings essential for treatment planning and assessment.
While previous research predominantly concentrated on a limited array of findings, such as tooth counting~\citep{oktay2017tooth, koch2019accurate}, missing teeth detection~\citep{tuzoff2019tooth, kim2020automatic, xu2023robust}, caries~\citep{dayi2023novel, zhu2022cariesnet, oztekin2023explainable, lian2021deep}, and oral lesions~\citep{endres2020development, yang2020deep}, recent studies have begun to explore a broader range of dental issues~\citep{muresan2020teeth, vinayahalingam2021automated, rohrer2022segmentation, almalki2022deep, bacsaran2022diagnostic}.
Some even tackled a more complex and challenging setting, similar to this study, where identification of findings and associating them with numbered teeth are both required~\citep{gardiyanouglu2023automatic,van2024combining}.
Nevertheless, these studies often suffer from limitations due to their reliance on data collected at single clinical sites~\citep{muresan2020teeth, rohrer2022segmentation, bacsaran2022diagnostic, lian2021deep, endres2020development, yang2020deep}, overlooking the variability in patient demographics and the discrepancies in image acquisition techniques across different setups, which crucially impact AI performance, especially when they are tasked to operate across different sites.
Moreover, many of these studies do not concurrently evaluate the performance of human readers~\citep{muresan2020teeth, rohrer2022segmentation, almalki2022deep, bacsaran2022diagnostic}, missing opportunities to demonstrate how AI systems could enhance support for dentist.

In this study, we examined the interpretation of DPRs using large multinational data sets.
We compiled three data sets encompassing a set of 8 dental finding categories to achieve a comprehensive evaluation of performance metrics.
Furthermore, we developed a novel AI system to concurrently perform finding identification and localization, and examined its generalizability across data sets from various continents.
Importantly, we also investigated the performance of human readers in interpreting DPRs under realistic conditions and provided essential insights into the potential integration of AI into clinical workflows to assist dentists and OMF surgeons.

%% file: includes/section-materials-and-methods.tex
\section{Materials and Methods}

\subsection{Data Sets}

\input{./includes/fig-data-flow}
\input{./includes/tbl-data-statistics}

This study analyzed $\num{6669}$ DPRs sourced from distinct clinical sites across three regions: the Netherlands, Brazil, and Taiwan (Figure~\ref{fig:data-flow}).
The sample size was determined based on the need to estimate the sensitivity and specificity metrics (>70\%) with a margin of error of ±3\% at a 95\% confidence level, leading to a minimum sample size of $\num{897}$.
We retrospectively collected data sets from the clinical sites, and further augmented radiographs from Brazil with a publicly available repository~\citep{silva2018automatic}.
Note for the aforementioned public data, we only utilized the radiograph images as their annotations did not align with the purpose of this study.

In the Netherlands, we included cases from various clinics in the Netherlands, acquired using several machines\footnote{\emph{Cranex Novus e} (Soredex, PaloDEx Group Oy, Finland), \emph{Orthopantomograph OP300} (Instrumentarium Dental, Finland), \emph{CS 8100 3D} (Carestream Dental LLC, United States), and \emph{PaX-i3D} (Vatech, Korea)}.
The data set from Brazil comprises \num{300} images sourced from \emph{\hospitalB}, and another \num{1500} images sourced from the open-access database.
As for Taiwan, we included cases from \emph{the Department of Dentistry at \hospitalT}\footnote{Obtained with a \emph{Veraviewepocs 2D} (J. Morita Mfg. Corporation., Japan) machine set at 60--80~kV, 1--10~mA, and 6.0--7.4~s, depending on the physical characteristics of the patients}.
All images were acquired using standard clinical protocols routinely employed at the respective hospitals.
We excluded DPRs from all sites that displayed mixed dentition, presence of metal accessories, or structural anomaly of bones (\eg, mandibular fractures or surgery thereof), as these conditions could potentially interfere with the imaging analysis.  
Additionally, images of inferior quality, which would typically necessitate a retake in a clinical setting, were also excluded.

The refined data sets comprised $\num{5245}$ images from the Netherlands, chronologically divided into $\num{4044}$ for AI training and validation, and $\num{1201}$ for AI testing and evaluation as the \emph{internal test set}. 
Also in the refined datasets were $\num{1173}$ images from Brazil and $\num{251}$ images from Taiwan, designated exclusively for AI testing and evaluation as the \emph{external test sets}.
Data collection procedures were reviewed and approved by the \emph{Institutional Review Board} of \hospitalN\footnote{Reference number: 2019-5232.} and \hospitalT\footnote{Reference number: \#202102018RINB.} respectively. 
All radiographs were anonymized and de-identified prior to their use in this study.

\subsection{Contour Labels for AI Training \& Reference for Evaluation}
\label{sec:label-for-training}

\input{./includes/fig-model-flow}

Two general dental practitioners both with 15 years of experience participated in the annotation of contour labels for AI training, as illustrated in Figure~\ref{fig:model-flow}. 
The training set comprised $\num{4044}$ DPRs from the Netherlands, analyzed for the 8 specific finding objects with no additional clinical information included:
\begin{enumerate*}[label=(\alph*)]
    \item presence of teeth,
    \item implants,
    \item residual roots, 
    \item crown/bridges, 
    \item root canal fillings, 
    \item fillings, 
    \item caries, and
    \item periapical radiolucencies.
\end{enumerate*}
Each object of interest was individually contoured by the first GDP, with concurrent annotation of tooth indices.
The annotated images, along with their associated contours and indices, were reviewed by the second GDP for accuracy, with adjustments made only where clearly necessary.
The annotated data set was then randomly divided into two subsets for AI system development: 70\% for training and 30\% for validation.

As for the evaluation of both AI and human reader assessment outcome, it was crucial to establish consistent reference finding assessments (\ie, the \emph{golden standard} for this study, as illustrated in Figure~\ref{fig:model-flow}) for each of the cases in the test sets.
The two GDPs who annotated the Netherlands training set also annotated the contour labels for the Netherlands test set and the Brazil test set similarly, and these contours were converted to the reference assessments.
For the Taiwan data, four GDPs (\gdpYW, \gdpCC, \gdpHC, and \gdpHW) were involved to produce the reference.
These four GDPs convened to standardize the diagnostic criteria for each finding prior to annotating their own assessments, and then they resolved discrepancies via a discussion to reach unanimous agreement for the final reference assessments.


\subsection{Human Reader Benchmarks for AI Comparison}

To establish a benchmark for our AI system, we randomly selected $\num{118}$ DPRs from the Taiwan data set (denoted as Taiwan*).
Four experienced dental practitioners were recruited as readers -- two general dental practitioners each with 2 and 3 years of experience (denoted as G1 and G2 in subsequent analyses), and two specialized dentists each with 11 years (prosthodontics/orthodontics) and 15 years (endodontics) of experience (denoted as S1 and S2).
These readers were distinct from the reference annotators mentioned in Section~\ref{sec:label-for-training}.
During the evaluation, all readers followed the same protocols established for the reference annotators for the Taiwan set.
However, unlike the reference annotations where a consensus was required, each benchmark reader independently submitted their own finding assessments.
These submissions were then individually evaluated against the reference.

\subsection{AI System}
\label{sec:ai-system}

The AI system employed in this study consists of a three-stage pipeline as depicted in Figure~\ref{fig:model-flow}.
Initially, an object detection model~\citep{redmon2016you} identifies and locates teeth presence and 7 other types of dental findings (as described in Section~\ref{sec:label-for-training}) within the DPRs.
The second stage involves classification of tooth indices at the pixel level using a semantic segmentation model~\citep{chen2017deeplab}.
Both stages utilize deep convolutional neural networks (CNNs), taking only the DPRs as input and no other information.
The results from these two stages are integrated during a post-processing stage, employing a probabilistic algorithm to correlate detected objects and classified indices, producing a finding assessment table for each case.
This table assigns a confidence score, ranging from zero to one, for each finding associated with each tooth, resulting in $\num{256}$ scores per DPR for evaluation.

AI system training and testing (case reading) was done on a workstation running on Ubuntu 22.04 with an Intel Core i9-10900 CPU, a nVIDIA GeForce RTX 3080 GPU, and 64GB of system memory.
The software\footnote{\url{https://github.com/stmharry/dental-pano-ai.git}.} was implemented using python 3.11 and various deep learning libraries.
Refer to Supplementary Material for a more detailed description on AI pipeline and training.

\subsection{Evaluation Metrics}


AI performance was evaluated using receiver operating characteristic (ROC) curves and the area under the ROC curve (AUC-ROC), estimated via the Wilcoxon \emph{U}-statistic with confidence intervals calculated by DeLong's method~\citep{delong1988comparing}. 
Key metrics, including sensitivity, specificity, precision, and F-scores, were computed with confidence intervals using Wald's method for binomial proportions.


Superiority and non-inferiority tests followed the Obuchowski-Rockette-Hillis (ORH) multi-reader multi-case (MRMC) framework, with jackknife resampling to estimate data covariances. 
Cohen's Kappa assessed agreement between human readers with CI calculated using McHugh's method~\citep{mchugh2012interrater}.
Statistical differences in Kappa distributions were analyzed via ORH methods.


Reading time comparisons included human readers and the AI system.
Evaluations for specific findings, such as \emph{implants}, were restricted to relevant tooth subsets (i.e., \emph{missing teeth} for implant evaluations and \emph{present teeth} for 6 other findings).
All metrics were calculated with 95\% confidence intervals (CI), and statistical significance was denoted for $p$-values below 0.05.


Analyses were conducted in Python 3.11 using MRMC-specific libraries~\citep{mckinney2022comparing}.

%% file: includes/fig-data-flow.tex
\begin{figure}[!t]
    \centering
    \includegraphics[width=1.0\textwidth, height=1.0\textheight, keepaspectratio]{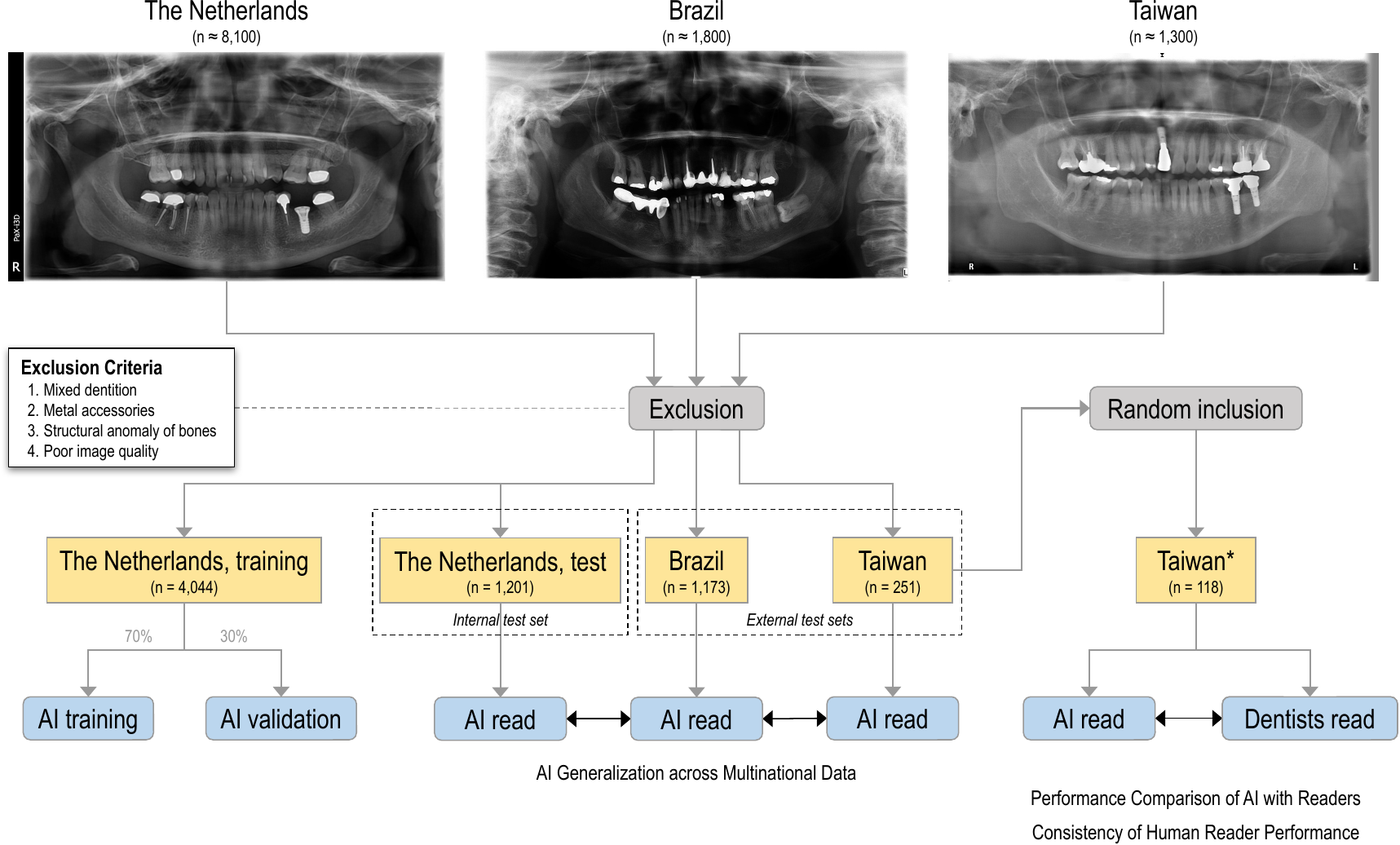}
    \caption{
        \textbf{Workflow for Dental Panoramic Radiograph Collection and Assessment Study.}
        Dental panoramic radiographs were collected from three geographic locations, totaling $\num{6669}$ studies: $\num{5245}$ from the Netherlands, $\num{1173}$ from Brazil, and $\num{251}$ from Taiwan.
        The initial data repositories before exclusion was much larger, with ``$\approx$'' denoting their original sizes.
        Data from the Netherlands were divided into a training set for artificial intelligence (AI) training and validation, and a test set designated as the \emph{internal test set}.
        Both the Brazil and Taiwan data sets were utilized exclusively as \emph{external test sets} for evaluation.
        The AI system was trained and validated on $\num{4044}$ imaging cases from the Netherlands, and was tasked with analyzing all three test sets to assess its generalizability.
        Additionally, both AI and dentists read a randomly selected subset of the Taiwan data to facilitate a comparative analysis of performance.
    }
    \label{fig:data-flow}
\end{figure}

%% file: includes/tbl-data-statistics.tex
\begin{table}[t]
\centering \caption{   \textbf{Characteristics of Patients in the Study.}   This table presents demographic and dental findings data from the test sets of the Netherlands, Brazil, and Taiwan.   The Netherlands test set is a chronologically divided subset of the larger data set from the Netherlands, which included the training set for AI.   Some demographic information, including the detailed age distribution from the Netherlands test set, and the sex/age from Brazil was not collected prior to anonymized analysis, and hence displayed as ``n.a.'', indicating that the statistics are not available.   \emph{Full dentition teeth count} refers to all possible teeth counts in the full dentition, or, 32 per study.   Counts for \emph{dental findings} represent the total occurrences of each finding across all teeth.   \emph{SD} stands for standard deviation, and \emph{IQR} denotes interquartile range, providing measures of variability and spread in the data, respectively. } \label{tbl:data-statistics} \small \rowcolors{1}{white}{table-bg}
\begin{tabular}{ p{0.35\linewidth}p{0.175\linewidth}p{0.175\linewidth}p{0.175\linewidth} }
\toprule
Characteristic & The Netherlands, test \newline ($n=\num{1201}$) & Brazil \newline ($n=\num{1173}$) & Taiwan \newline ($n=\num{251}$) \\ \midrule
Years & n.a. & n.a. & 2021 \\ \midrule
Sex, count (\%) &   &   &   \\
{~~~~}Female & 558 (46.5\%) & n.a. & 143 (56.7\%)  \\
{~~~~}Male & 643 (53.5\%) & n.a. & 109 (43.3\%) \\ \midrule
Age, years &   &   &   \\
{~~~~}Mean (SD) & 36.8 (n.a.) & n.a. & 56.4 (19.4) \\
{~~~~}Median (IQR) & n.a. & n.a. & 58.5 (45.0--71.8) \\
{~~~~}Range & 16--88 & n.a. & 18--95 \\ \midrule
Dental findings, count (prevalence \%) &   &   &   \\
{~~~~}Full dentition teeth count & \num{38432} (100.00\%) & \num{37536} (100.00\%) & \num{8032} (100.00\%) \\
{~~~~}Missing & \num{6836} (17.79\%) & \num{5926} (15.79\%) & \num{1819} (22.65\%) \\
{~~~~}Implant & \num{621} (1.62\%) & \num{586} (1.56\%) & \num{198} (2.47\%) \\
{~~~~}Residual root & \num{138} (0.36\%) & \num{200} (0.53\%) & \num{89} (1.11\%) \\
{~~~~}Crown/bridge & \num{3865} (10.06\%) & \num{1419} (3.78\%) & \num{1004} (12.50\%) \\
{~~~~}Root canal filling & \num{1867} (4.86\%) & \num{1317} (3.51\%) & \num{648} (8.07\%) \\
{~~~~}Filling & \num{7412} (19.29\%) & \num{8144} (21.70\%) & \num{1370} (17.06\%) \\
{~~~~}Caries & \num{1063} (2.77\%) & \num{1442} (3.84\%) & \num{317} (3.95\%) \\
{~~~~}Periapical radiolucency & \num{458} (1.19\%) & \num{380} (1.01\%) & \num{197} (2.45\%) \\ \bottomrule
\end{tabular}
\end{table}

%% file: includes/fig-model-flow.tex
\begin{figure}[!t]
    \centering
    \includegraphics[width=1.0\textwidth, height=1.0\textheight, keepaspectratio]{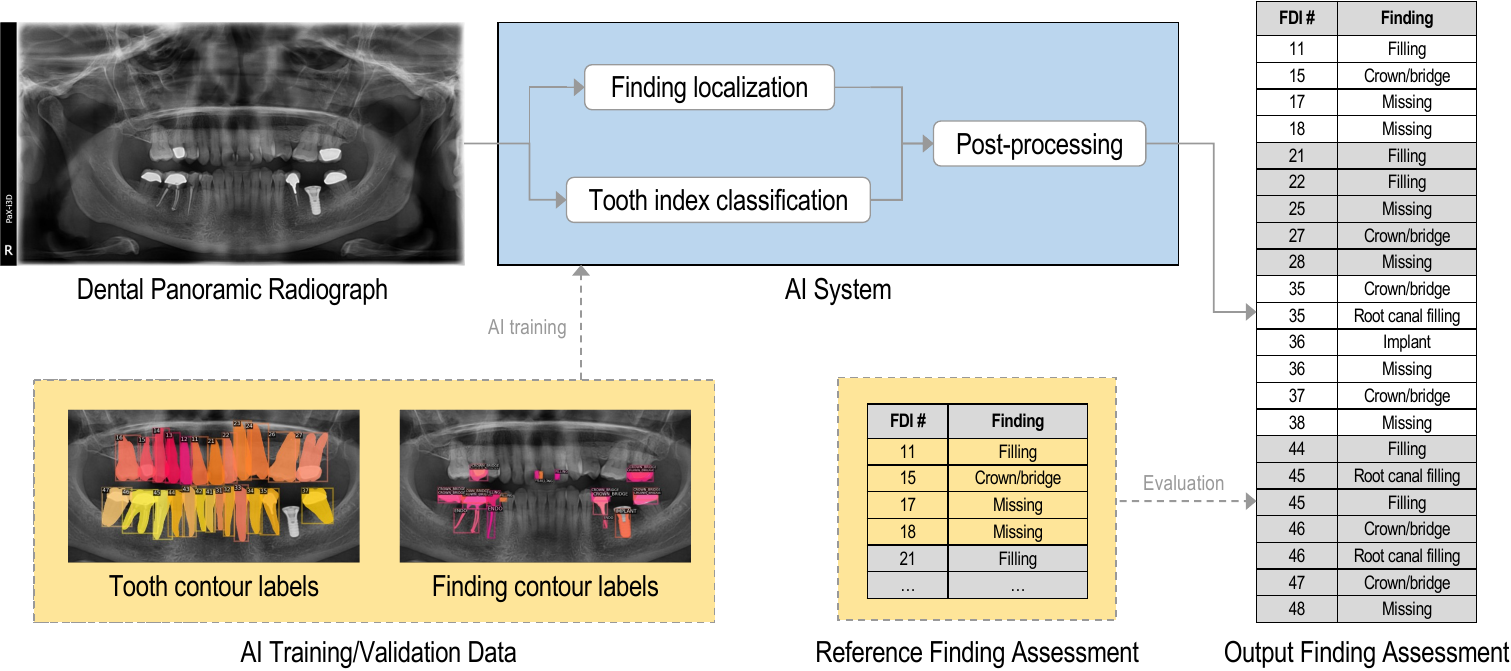}
    \caption{
        \textbf{Overview of the Data Modalities and Artificial Intelligence (AI) Workflow.}
        This schematic illustrates the AI process flow, beginning with dental panoramic radiographs (DPRs) as the primary input and culminating in a detailed finding assessment that labels whether each of the 8 findings is present in each of the 32 teeth (hence 256 binary labels per image).
        The AI system operated through three main stages: dental finding localization, tooth index classification, and post-processing.
        For AI training, we utilized contour labels annotated by general dental practitioners on the Netherlands training set.
        Note that only positive finding labels are shown in the finding assessment in the workflow, and for each positive finding, it is indicated by a finding type and a tooth number, which was annotated using the \emph{FDI World Dental Federation} notation, or, \emph{ISO-3950} notation in this study.
        The FDI notation categorizes teeth into four quadrants, each having eight teeth, resulting in a number ranging from 11 to 48.
    }
    \label{fig:model-flow}
\end{figure}

%% file: includes/section-results.tex
\section{Results}

\subsection{Data Set Characteristics}
The data sets encompassed $\num{6669}$ patients, each represented by one dental panoramic radiograph (DPR), across continents and clinical sites. 
The training data set (from the Netherlands) comprised $\num{4044}$ patients, while the test data sets (from the Netherlands, Brazil, Taiwan) included a total of $\num{2625}$ patients (48.2\% female and 51.7\% male\footnote{Accounting only the Netherlands test set and the Taiwan set, where demographics were known.}).
The overall mean age was 40.2 years.
The test DPRs contained a total of \num{84000} full dentition teeth slots, exhibiting various findings.
The prevalence of findings included fillings ($n=\num{16926}$; 20.15\%), missing teeth ($n=\num{14581}$; 17.36\%), crown/bridges ($n=\num{6288}$; 7.49\%), root canal fillings ($n=\num{3832}$; 4.56\%), caries ($n=\num{2822}$; 3.36\%), implants ($n=\num{1405}$; 1.67\%), periapical radiolucencies ($n=\num{1035}$; 1.23\%), and residual roots ($n=\num{427}$; 0.51\%).
Detailed statistics of the data sets are provided in Table~\ref{tbl:data-statistics}.

\subsection{Performance Comparison of AI with Human Readers}
\label{sec:ai-vs-reader}

\input{./includes/fig-ai-vs-reader}

In the Taiwan* test subset of $\num{118}$ images, we assessed the performance of the AI system and four readers (G1, G2, S1, and S2) against the reference.
The AI system's receiver operating characteristic (ROC) curve and the set operating points are displayed in Figure~\ref{fig:ai-vs-reader}.
Supplementary Material includes more numerical results.
All operating points were chosen for the optimal $\textrm{F}_2$ score to prioritize sensitivity for screening scenarios.
AUC-ROC values for all findings exceeded 80\%, reaching a macro-averaged AUC-ROC of 96.2\% (95\% CI: 94.6\%--97.8\%) across 8 finding types.

The AI illustrated a statistically significant improvement in sensitivity for periapical radiolucencies at +67.9\% (95\% CI: 54.0\%--81.9\%; $p < .001$ for superiority) compared to the average human reader, while maintaining non-inferiority in specificity ($p < .001$ at a pre-specified 5\% margin).
Similarly, for missing teeth, the AI system demonstrated, over the average reader, a statistically significant improvement in sensitivity at +4.7\% (95\% CI: 1.4\%--8.0\%; $p = .008$) along with non-inferiority in specificity ($p < .001$).

For other dental findings, the AI system achieved simultaneous non-inferiority in both sensitivity ($p = .0049$) and specificity ($p < .001$) for implants, crown/bridges ($p = .0012$, $p < .001$), root canal fillings ($p = .0090$, $p < .001$), and caries ($p = .040$, $p < .001$).
Specificity for residual roots showed non-inferiority ($p < .001$), and for fillings, sensitivity also demonstrated non-inferiority ($p = .034$).

\subsection{AI Generalization across Multinational Data}
\label{sec:ai-generalizability}

\input{./includes/tbl-ai-multinational-short}
\input{./includes/fig-ai-generalizability}

The AI system, trained exclusively on the Netherlands training set, was tested and evaluated on the Netherlands test set, the Brazil set, and the Taiwan set.
An operating point optimized for screening settings was chosen similarly to Section~\ref{sec:ai-vs-reader}, and the system's performance was compared against the reference finding summary, as depicted in Figure~\ref{fig:ai-generalizability} and Table~\ref{tbl:ai-multinational-short}.

Across both internal (the Netherlands) and external (Brazil and Taiwan) test sets, the AUC-ROC scores were consistently above 80\% for all findings.
The macro-averaged AUC-ROC across test sets was 99.4\% (95\% CI: 99.1\%--99.7\%) for root canal fillings, 99.2\% (95\% CI: 98.3\%--100.0\%) for implants, and 98.8\% (95\% CI: 97.7\%--99.9\%) for crown/bridges.
AUC-ROC scores for periapical radiolucencies and caries were 92.1\% (95\% CI: 87.7\%--96.6\%) and 81.5\% (95\% CI: 79.1\%--84.0\%), respectively.

The average sensitivity across test sets was highest for implants at 97.6\% (95\% CI: 95.0\%--100.0\%) and lowest for caries at 52.1\% (95\% CI: 44.7\%--59.5\%).
Similarly, specificity peaked for residual roots at 99.5\% (95\% CI: 99.0\%--100.0\%) and was lowest for caries at 90.4\% (95\% CI: 84.6\%--96.1\%).
Precision was highest for root canal fillings at 91.4\% (95\% CI: 86.6\%--96.2\%) and notably lower for caries at 21.9\% (95\% CI: 9.7\%--34.1\%).

Evaluating discrepancies in AI performance between the internal and external test sets revealed no statistically significant differences in AUC-ROC or sensitivity across all findings.
However, specificity for root canal fillings showed a statistically significant decrease in the external test sets compared to the internal test set ($p = .035$).
Similarly, precision for residual roots was significantly lower in the external test sets ($p = .042$).

As for the agreements between the AI and the reference, see Supplementary Material for a detailed discussion.

\subsection{Reading Times}

The average reading time on a DPR for the four human readers was 122 seconds (95\% CI: 118~s--126~s; IQR: 79~s--155~s).
In comparison, the AI system demonstrated markedly faster processing times across its three stages.
The first stage, which involved detecting findings in the DPRs, required an average of 0.28 seconds per case (95\% CI: 0.27~s--0.28~s; IQR: 0.26~s--0.30~s).
The second stage, dedicated to classifying tooth indices, took 0.20 seconds (95\% CI: 0.19~s--0.20~s; IQR: 0.18~s--0.22~s).
The final post-processing stage required 1.08 seconds (95\% CI: 1.05~s--1.11~s; IQR: 1.00~s--1.14~s).
Consequently, the total runtime for the AI system averaged 1.55 seconds per image (95\% CI: 1.52~s--1.58~s; IQR: 1.43~s--1.66~s), based on the machine specifications detailed in Section~\ref{sec:ai-system}.

%% file: includes/fig-ai-vs-reader.tex
\begin{figure}[!t]
    \centering
    \includegraphics[width=1.0\textwidth, height=1.0\textheight, keepaspectratio]{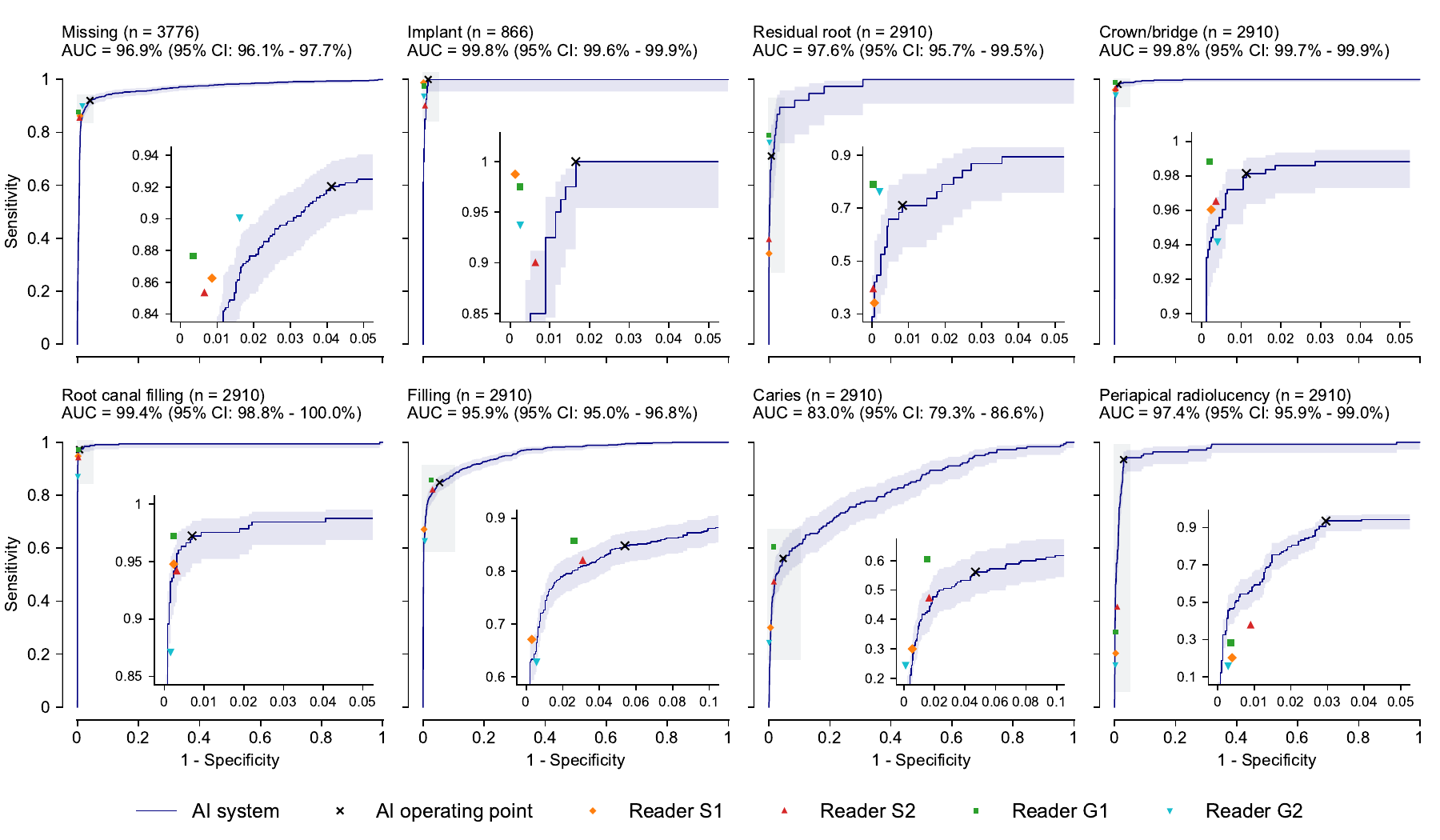}
    \caption{
        \textbf{Comparison of AI System Performance with Human Readers on Dental Finding Assessment in Taiwan.}
        This figure presents the receiver operating characteristic (ROC) curves for the AI system alongside the performance of 4 human readers on the Taiwan* test subset.
        Each plot displays a specific dental finding, with the shaded vertical areas representing the 95\% confidence intervals (CI) for sensitivity along the ROC curve.
        The AI system’s operating point, indicated by ``×'', was determined by maximizing the $\textrm{F}_2$ score on a held-out validation set, tailored for screening scenarios.
        Figure insets magnify the critical regions of interest within each graph, providing a detailed view of performance near the operating point.
    }
    \label{fig:ai-vs-reader}
\end{figure}

%% file: includes/tbl-ai-multinational-short.tex
\begin{table}[!h]
\centering\caption{  \textbf{Performance of AI on Multiple Data Sets.}   This table provides a horizontal comparison of various performance metrics for the AI system across internal and external test sets.   The rightmost columns present macro-averages of the metrics across either external test sets or all test sets.  For the mean external metrics, discrepancy tests were conducted against internal performance metrics, with ``n.s.'' indicating no significant difference and ``*'' denoting $p<0.05$.  The AI was optimized to maximize the $\textrm{F}_2$ score for a screening setting.  ``--'' indicates that the metric is not applicable.  Values in parentheses represent 95\% confidence intervals. }\label{tbl:ai-multinational-short}\scriptsize\rowcolors{1}{white}{table-bg}
\begin{tabular}{ p{0.2\linewidth}p{0.15\linewidth}p{0.125\linewidth}p{0.125\linewidth}|p{0.15\linewidth}p{0.125\linewidth} }
\toprule
Data Set & The Netherlands, test & Brazil & Taiwan & Mean, external [v.s. internal] & Mean, overall \\
Test Set Type & Internal & External & External & External &   \\ \midrule
Missing &   &   &   &   &   \\
{~~~~}Positive/negative count & \num{6836} / \num{31596} & \num{5926} / \num{31610} & \num{1819} / \num{6213} & -- & -- \\
{~~~~}Metric, \% (95\% CI) &   &   &   &   &   \\
{~~~~~~~~}Sensitivity & 87.6 {(86.8--88.4)} & 93.6 {(92.9--94.2)} & 89.3 {(87.8--90.7)} & 91.5 (87.2--95.7) [n.s.] & 90.2 (86.6--93.8) \\
{~~~~~~~~}Specificity & 95.5 {(95.3--95.8)} & 95.7 {(95.5--95.9)} & 95.1 {(94.5--95.6)} & 95.4 (94.7--96.1) [n.s.] & 95.4 (94.9--95.9) \\
{~~~~~~~~}Precision & 80.9 {(80.0--81.8)} & 80.3 {(79.3--81.2)} & 84.2 {(82.5--85.7)} & 82.2 (78.2--86.2) [n.s.] & 81.8 (79.1--84.4) \\
{~~~~~~~~}AUC-ROC & 94.4 {(94.0--94.9)} & 97.8 {(97.6--98.1)} & 95.7 {(95.1--96.4)} & 96.8 (94.7--98.9) [n.s.] & 96.0 (94.0--98.0) \\ \midrule

Implant &   &   &   &   &   \\
{~~~~}Positive/negative count & \num{619} / \num{6217} & \num{586} / \num{5340} & \num{191} / \num{1628} & -- & -- \\
{~~~~}Metric, \% (95\% CI) &   &   &   &   &   \\
{~~~~~~~~}Sensitivity & 98.2 {(96.8--99.0)} & 95.6 {(93.6--97.0)} & 99.0 {(96.3--99.7)} & 97.3 (93.5--100.0) [n.s.] & 97.6 (95.0--100.0) \\
{~~~~~~~~}Specificity & 97.6 {(97.2--98.0)} & 95.0 {(94.4--95.6)} & 98.3 {(97.5--98.8)} & 96.6 (93.4--99.9) [n.s.] & 97.0 (94.9--99.0) \\
{~~~~~~~~}Precision & 80.3 {(77.3--83.0)} & 67.8 {(64.5--70.9)} & 87.1 {(82.0--90.9)} & 77.4 (58.1--96.8) [n.s.] & 78.4 (66.8--90.0) \\
{~~~~~~~~}AUC-ROC & 99.5 {(99.4--99.6)} & 98.3 {(98.1--98.6)} & 99.7 {(99.6--99.9)} & 99.0 (97.7--100.0) [n.s.] & 99.2 (98.3--100.0) \\ \midrule

Residual root &   &   &   &   &   \\
{~~~~}Positive/negative count & \num{137} / \num{31459} & \num{198} / \num{31412} & \num{88} / \num{6125} & -- & -- \\
{~~~~}Metric, \% (95\% CI) &   &   &   &   &   \\
{~~~~~~~~}Sensitivity & 79.6 {(72.0--85.5)} & 56.1 {(49.1--62.8)} & 72.7 {(62.6--80.9)} & 64.4 (46.2--82.6) [n.s.] & 69.4 (53.8--85.1) \\
{~~~~~~~~}Specificity & 99.9 {(99.8--99.9)} & 99.5 {(99.4--99.6)} & 99.0 {(98.8--99.3)} & 99.3 (98.8--99.8) [n.s.] & 99.5 (99.0--100.0) \\
{~~~~~~~~}Precision & 76.2 {(68.6--82.5)} & 41.7 {(36.0--47.7)} & 52.0 {(43.3--60.7)} & 46.9 (34.3--59.4) [*] & 56.7 (35.4--78.0) \\
{~~~~~~~~}AUC-ROC & 93.1 {(89.8--96.4)} & 92.8 {(90.3--95.3)} & 98.3 {(97.4--99.2)} & 95.5 (89.8--100.0) [n.s.] & 94.7 (90.5--99.0) \\ \midrule

Crown/bridge &   &   &   &   &   \\
{~~~~}Positive/negative count & \num{3038} / \num{28558} & \num{896} / \num{30714} & \num{851} / \num{5362} & -- & -- \\
{~~~~}Metric, \% (95\% CI) &   &   &   &   &   \\
{~~~~~~~~}Sensitivity & 92.7 {(91.7--93.5)} & 87.4 {(85.1--89.4)} & 97.2 {(95.8--98.1)} & 92.3 (82.5--100.0) [n.s.] & 92.4 (86.7--98.2) \\
{~~~~~~~~}Specificity & 98.7 {(98.6--98.8)} & 98.8 {(98.7--98.9)} & 99.4 {(99.1--99.6)} & 99.1 (98.5--99.7) [n.s.] & 99.0 (98.5--99.4) \\
{~~~~~~~~}Precision & 88.5 {(87.4--89.6)} & 68.3 {(65.5--70.9)} & 96.2 {(94.7--97.3)} & 82.2 (54.8--100.0) [n.s.] & 84.3 (67.9--100.0) \\
{~~~~~~~~}AUC-ROC & 98.7 {(98.5--98.9)} & 98.0 {(97.4--98.6)} & 99.8 {(99.7--99.9)} & 98.9 (97.1--100.0) [n.s.] & 98.8 (97.7--99.9) \\ \midrule

Root canal filling &   &   &   &   &   \\
{~~~~}Positive/negative count & \num{1867} / \num{29729} & \num{1316} / \num{30294} & \num{643} / \num{5570} & -- & -- \\
{~~~~}Metric, \% (95\% CI) &   &   &   &   &   \\
{~~~~~~~~}Sensitivity & 97.4 {(96.5--98.0)} & 94.1 {(92.7--95.3)} & 96.7 {(95.1--97.9)} & 95.4 (92.6--98.3) [n.s.] & 96.1 (93.8--98.3) \\
{~~~~~~~~}Specificity & 99.6 {(99.5--99.7)} & 99.4 {(99.3--99.5)} & 99.3 {(99.0--99.5)} & 99.3 (99.1--99.5) [*] & 99.4 (99.2--99.6) \\
{~~~~~~~~}Precision & 93.6 {(92.4--94.6)} & 86.8 {(84.9--88.4)} & 93.8 {(91.7--95.4)} & 90.3 (83.1--97.4) [n.s.] & 91.4 (86.6--96.2) \\
{~~~~~~~~}AUC-ROC & 99.5 {(99.3--99.7)} & 99.2 {(99.0--99.5)} & 99.5 {(99.1--99.8)} & 99.3 (98.9--99.7) [n.s.] & 99.4 (99.1--99.7) \\ \midrule

Filling &   &   &   &   &   \\
{~~~~}Positive/negative count & \num{7412} / \num{24184} & \num{8127} / \num{23483} & \num{1370} / \num{4843} & -- & -- \\
{~~~~}Metric, \% (95\% CI) &   &   &   &   &   \\
{~~~~~~~~}Sensitivity & 92.2 {(91.6--92.8)} & 93.8 {(93.3--94.3)} & 87.6 {(85.7--89.2)} & 90.7 (84.5--96.9) [n.s.] & 91.2 (87.4--95.0) \\
{~~~~~~~~}Specificity & 93.9 {(93.6--94.2)} & 92.3 {(91.9--92.6)} & 88.0 {(87.1--88.9)} & 90.2 (85.9--94.4) [n.s.] & 91.4 (87.9--94.9) \\
{~~~~~~~~}Precision & 82.3 {(81.5--83.1)} & 80.8 {(80.0--81.6)} & 67.4 {(65.2--69.6)} & 74.1 (60.9--87.3) [n.s.] & 76.8 (67.5--86.2) \\
{~~~~~~~~}AUC-ROC & 97.3 {(97.0--97.5)} & 97.2 {(97.0--97.4)} & 94.8 {(94.1--95.5)} & 96.0 (93.5--98.5) [n.s.] & 96.4 (94.7--98.1) \\ \midrule

Caries &   &   &   &   &   \\
{~~~~}Positive/negative count & \num{1063} / \num{30533} & \num{1435} / \num{30175} & \num{317} / \num{5896} & -- & -- \\
{~~~~}Metric, \% (95\% CI) &   &   &   &   &   \\
{~~~~~~~~}Sensitivity & 46.3 {(43.3--49.3)} & 57.4 {(54.8--59.9)} & 52.7 {(47.2--58.1)} & 55.0 (48.8--61.3) [n.s.] & 52.1 (44.7--59.5) \\
{~~~~~~~~}Specificity & 91.8 {(91.5--92.1)} & 84.7 {(84.3--85.1)} & 94.5 {(93.9--95.1)} & 89.6 (80.0--99.2) [n.s.] & 90.4 (84.6--96.1) \\
{~~~~~~~~}Precision & 16.5 {(15.2--17.9)} & 15.2 {(14.2--16.1)} & 34.1 {(30.0--38.4)} & 24.6 (5.8--43.4) [n.s.] & 21.9 (9.7--34.1) \\
{~~~~~~~~}AUC-ROC & 82.5 {(81.3--83.7)} & 79.8 {(78.6--81.0)} & 82.3 {(79.6--84.9)} & 81.0 (77.9--84.2) [n.s.] & 81.5 (79.1--84.0) \\ \midrule

Periapical radiolucency &   &   &   &   &   \\
{~~~~}Positive/negative count & \num{458} / \num{31138} & \num{377} / \num{31233} & \num{197} / \num{6016} & -- & -- \\
{~~~~}Metric, \% (95\% CI) &   &   &   &   &   \\
{~~~~~~~~}Sensitivity & 55.2 {(50.7--59.7)} & 53.8 {(48.8--58.8)} & 86.8 {(81.4--90.8)} & 70.3 (37.7--100.0) [n.s.] & 65.3 (43.7--86.9) \\
{~~~~~~~~}Specificity & 98.2 {(98.1--98.3)} & 97.2 {(97.1--97.4)} & 94.9 {(94.3--95.4)} & 96.1 (93.7--98.4) [n.s.] & 96.8 (94.8--98.7) \\
{~~~~~~~~}Precision & 31.2 {(28.1--34.4)} & 19.1 {(16.9--21.6)} & 35.8 {(31.6--40.2)} & 27.4 (10.8--44.1) [n.s.] & 28.7 (18.4--39.0) \\
{~~~~~~~~}AUC-ROC & 93.2 {(92.0--94.5)} & 88.1 {(85.9--90.3)} & 95.0 {(93.3--96.8)} & 91.6 (84.5--98.6) [n.s.] & 92.1 (87.7--96.6) \\ \bottomrule
\end{tabular}
\end{table}

%% file: includes/fig-ai-generalizability.tex
\begin{figure}[!t]
    \centering
    \includegraphics[width=1.0\textwidth, height=0.8\textheight, keepaspectratio]{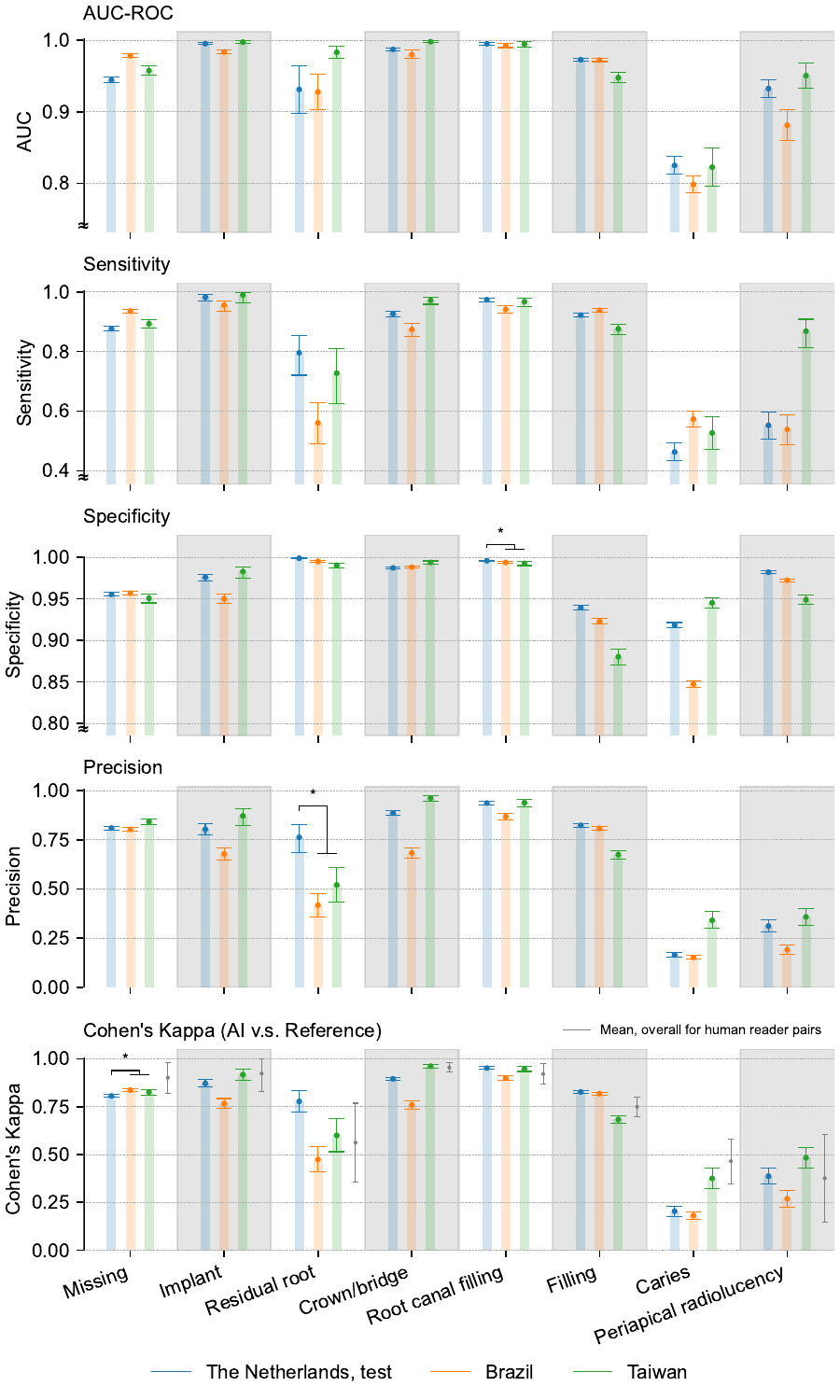}
    \caption{
        \textbf{AI Generalization Performance across Multinational Data Sets.}
        This figure evaluates the AI's capability to generalize its performance across different geographic data sets, focusing on the assessment of DPRs.
        The operating point of the AI system was optimized to maximize the $\textrm{F}_2$ score on a held-out validation set, simulating a screening scenario.
        Each bar represents the AI's performance metric for a specific dental finding within a dataset, with the 95\% confidence intervals shown as error bars.
        Notably, the $y$-axes for some metrics do not start from zero to highlight specific performance ranges.
        Cohen's Kappa values among pairs of the human readers (G1, G2, S1, and S2) were computed and displayed along the AI's Kappa against the reference, serving as a contextual upper limit for AI performance.
        A comprehensive exploration on the inter-human-reader agreements is included in Supplementary Material.
    }
    \label{fig:ai-generalizability}
\end{figure}

%% file: includes/section-discussion.tex
\section{Discussion}

Assessing DPRs involves identifying findings and accurately localizing them relative to anatomical landmarks, a dual challenge not addressed by most previous studies. 
To our knowledge, this study is the first to establish a comprehensive benchmark for AI systems on DPR assessment, demonstrating performance comparable to human readers and excelling in specific findings.



Most previous works employed \emph{detection-only} metrics for dental finding evaluation, which overlook correspondence to FDI-labeled tooth numbering. 
For instance, in detecting missing teeth, \citet{tuzoff2019tooth} reported sensitivity/specificity of 99.41\%/99.45\%, \citet{muramatsu2021tooth} achieved a sensitivity of 96.4\%, and \citet{leite2021artificial} attained sensitivity/precision of 98.9\%/99.6\%.
For works tackling the dual challenge, we reported sensitivity/specificity of 90.2\%/95.4\%, surpassing the 75.5\% sensitivity reported by \citet{kim2020automatic} and comparable to the 96.5\% reported by \citet{vinayahalingam2021automated}.
Furthermore, our AI’s specificity was 95.9\% (95\% CI: 95.1\%--96.5\%), exceeding the 80.4\% from \citet{kim2020automatic} and comparable to the 97.2\% reported by \citet{chen2022missing}.

A limited number of studies have concurrently analyzed multiple findings in DPRs~\citep{vinayahalingam2021automated, bacsaran2022diagnostic, van2024combining}.
Our AI system demonstrated exceptional performance across several metrics.
For root canal fillings, our $\textrm{F}_1$ score reached 93.7\% (95\% CI: 90.2\%--97.1\%), surpassing prior works (81.96\%--88.6\%).
For implants, our $\textrm{F}_1$ score of 86.8\% (95\% CI: 78.7\%--94.8\%) was comparable to reported values of 80.9\%--94.33\%. Similarly, for fillings, our $\textrm{F}_1$ score of 83.3\% (95\% CI: 76.2\%--90.4\%) aligned with previous results (83.0\%--86.84\%).

However, directly comparing performance across studies is inherently complex, as such comparisons disregard variations in data distributions, collection processes, curation methodologies, and AI operating point selection.
As \citet{van2024combining} proposed, combining public DPR datasets into a unified benchmarking set can help address this challenge.
Nevertheless, practical research often prioritizes novel directions, such as exploring new clinical findings, imaging modalities, or diversified test data, making standardized data consolidation impractical.

To address these challenges, we propose evaluating AI performance relative to inter-human-reader agreement (\emph{human agreement}) levels among human experts, measured using overall mean of Cohen's Kappa between any pair of human readers.
These agreements represent an approximate upper bound for AI performance, as even diagnostic consensus among clinical experts rarely exceeds such values. 
Our study revealed minimal to moderate human agreement levels for residual roots (56.2\%, 95\% CI: 35.6\%--76.8\%), fillings (74.9\%, 95\% CI: 69.8\%--80.0\%), caries (46.5\%, 95\% CI: 34.8\%--58.1\%), and periapical radiolucencies (37.6\%, 95\% CI: 14.7\%--60.6\%). 
These results underscore the inherent clinical uncertainty that persists despite predefined diagnostic standards. 
Notably, our AI system achieved Kappa values comparable to human agreement across most findings, with the exception of caries detection ($p = .024$, two-sample $t$-test), indicating room for further improvement.


Another critical consideration often overlooked in prior studies is the need for diverse, adequately sized external test sets.
Most studies utilized single-site datasets; for example, \citet{van2024combining} used two datasets (458 cases and 4 findings), while \citet{bacsaran2022diagnostic} analyzed 10 findings but with only 118 cases.
Our study is, to the best of our knowledge, the most comprehensive to date, incorporating three test sets---two of which were \emph{external}---and including a total of $\num{2625}$ cases spanning 8 types of findings.

Despite being trained on data from the Netherlands, our AI system demonstrated strong generalizability on external test sets from Brazil and Taiwan, each with distinct imaging characteristics and treatment preferences.
For example, the Netherlands dataset included textual labels, the Brazil dataset exhibited strong contrast and spinal cord inclusion, and the Taiwan dataset displayed higher overall brightness.
Regionally, treatment preferences also varied, such as Taiwan's higher third molar extraction rate (8.6\%)~\citep{liang2022tooth}, resulting in elevated prevalence of missing teeth and residual roots.



This study is not without limitations.
The training data annotations did not fully adhere to the diagnostic standards established herein, particularly for findings with low prevalence (\eg, caries, residual roots, and periapical radiolucencies).
While this did not affect the validity of our claims, it highlights room for improvement, as high-quality training data often drive the last bit of gains when already using state-of-the-art models.
Additionally, our reader group, though diverse, lacked representation from some dental specialties and international training backgrounds, which could have enriched comparisons with AI performance~\citep{endres2020development}.


The optimal integration of AI into clinical workflows remains an open question, as this study did not assess active deployment scenarios.
However, the high sensitivity and adjustable operating points of our AI system suggest its potential to augment clinical practice, particularly for findings that may be overlooked due to time constraints or human error.

In conclusion, we implemented an AI system for detecting findings in dental panoramic radiographs (DPRs) with performance comparable to human readers for 7 of 8 included findings.
The system was validated across three continents, highlighting real-world challenges and opportunities for AI integration into dental practice.
Its time efficiency and operational benefits hold significant potential for enhancing clinical workflows.

%% file: includes/section-supplementary-material.tex
\section{AI System Pipeline and Training}
\label{sec:ai-system-pipeline}

In this study, our AI system followed a three-stage pipeline:
\begin{enumerate*}[label=(\alph*)]
    \item the finding localization module,
    \item the tooth index classification module, and
    \item the post-processing module.
\end{enumerate*}
The first two modules utilized convolutional neural networks (CNNs), while the post-processing module employed a probabilistic approach to integrate the outputs from the prior modules and produce the final AI assessment scores.

\subsection{Finding Localization Module}
The finding localization problem was formulated as an \emph{object detection} task in computer vision. 
We employed the YOLOv8 model~\citep{varghese2024yolov8}, a state-of-the-art CNN architecture for object detection.
YOLOv8 improved upon its predecessors with enhanced efficiency and accuracy, supporting high-resolution inputs (1024×1024) and complex segmentation tasks, making it particularly suitable for medical image analysis.
Specifically, we used the medium variant of the YOLOv8 model family.

The model was initialized with pre-trained weights from the Common Objects in Context (COCO) challenge~\citep{lin2014microsoft}.
Training was conducted on the Netherlands data of \num{4044} DPRs, split into 70\% for training and 30\% for validation.
The training spanned up to \num{50} epochs, using a batch size of 1 and an image size of 1024×1024 to retain high-resolution details essential for accurate localization.
A cosine decay learning rate schedule was applied, starting at \num{0.01} and decreasing gradually to stabilize performance.

To accelerate convergence, a warm-up phase of 3 epochs linearly increased the learning rate.
Momentum was set to 0.937 for a balance between stability and speed, while weight decay at 0.0005 prevented overfitting.
Data augmentation strategies, including random scaling (0.5), translation (0.1), and rotation (15 degrees), were applied to improve generalization.

During testing time (\ie, inference, or, case reading), the module generated a list of candidate objects. Each object instance $i$ included:
\begin{enumerate}
    \item an objectness score $\objness_i\in\zeroOne$;
    \item an finding probability vector $\objProb_i\in\zeroOne^C$, where $C=8$ is the number of finding types;
    \item a bounding box; and
    \item a probabilistic contour (mask) $\objMap_i\in\zeroOne^{H \times W}$, where $H$ and $W$ denote the image height and width. 
\end{enumerate}

The object list was sorted in descending order of objectness, and non-maximum suppression was applied to remove overlapping objects exceeding 50\% area overlap.
A maximum of \num{300} objects with objectness scores as low as \num{0.0001} were retained.

\subsection{Tooth Index Classification Module}
The tooth index classification module aimed to associate each image pixel with a specific FDI (World Dental Federation) index using a \emph{semantic segmentation} approach.
This module employed the DeepLabv3 model~\citep{chen2017rethinking}, which integrates a 101-layer backbone for feature extraction and an Atrous Spatial Pyramid Pooling (ASPP) module to capture multi-scale contextual information via atrous convolutions at varying rates.

We implemented the model using Detectron2~\citep{wu2019detectron2}, enabling efficient large-scale segmentation.
Input images were resized to 512×512, with dilated convolutions applied at rates of 6, 12, and 18 to expand the receptive field while preserving spatial resolution.
High-resolution low-level features from earlier backbone layers were projected via the DeepLabv3 head configuration for fine-grained segmentation.

Contours corresponding to the 32 full-dentition teeth were extracted and flattened onto maps matching the input image dimensions.
The model was trained for \num{100000} steps with a batch size of 3 images.
A linear learning rate warmup phase over the first \num{1000} iterations ensured stable optimization, after which the learning rate decayed via a cosine schedule to near-zero values.
A base learning rate of \num{0.001} was used, combined with momentum (0.9) and weight decay (\num{0.0001}) to reduce overfitting.
Gradient clipping at 1.0 prevented exploding gradients.
Each training iteration processed input images resized dynamically between 768×1024 and 1280×2048 to introduce variability and enhance generalization.

During inference, the module produced a dense probability map with 33 channels (32 tooth classes and 1 background class), denoted as $\semMap_j\in\zeroOne^{H \times W}$, where $j$ runs over $J=33$ classes, and $H$ and $W$ are the image dimensions.

\subsection{Post-Processing Module}
The post-processing module integrated outputs from the finding localization and tooth index classification modules.
Specifically, we obtained:
\begin{equation}
    \paral{\objness_i, \objProb_i, \objMap_i}_{i=1}^I \quad \text{and} \quad \paral{\semMap_j}_{j=1}^J.
\end{equation}

The correlation $\corr_{ij}$ between a finding object (instance) $i$ and a tooth index (semantic class) $j$ was defined as:
\begin{equation}
    \corr_{ij} \equiv \frac{\paraa{\objMap_i, \semMap_j}}{\sqrt{\paraa{\objMap_i, \objMap_i}}} \in \zeroOne,
\end{equation}
where $\paraa{\cdot, \cdot}$ denotes the inner product of probability maps, summing over the pixel-wise probability products over the whole image.

Following~\citet{ardila2019end}, we applied a \emph{soft-or} strategy to aggregate the contributions of all objects $i$ with detection scores $\objness_i$ and correlations $\corr_{ij}$.
The presence probability of finding $c$ in tooth index $j$, denoted $\para{\semProb_j}_c$, was calculated as:
\begin{equation}
    1 - \para{\semProb_j}_c \equiv \prod_i \param{ 1 - \func_c\para{\objness_i, \corr_{ij}; \paral{r}} }^{\para{\objProb_i}_c},
\end{equation}
where $\paral{r}$ is the set of all correlations, and $\func_c$ is a finding-dependent function defined as:
\begin{equation}
    \func_c\para{s_i, r_{ij}; \paral{r}} \equiv
    \begin{cases}
        \frac{r_{ij}}{\sum_{j\neq\text{background}} r_{ij}}, & \text{if } c = \text{missing}; \\
        s_i \cdot \frac{r_{ij}}{\sum_{j\neq\text{background}} r_{ij}}, & \text{for other findings}.
    \end{cases}
\end{equation}

The final output consisted of a set of probabilities:
\begin{equation}
    \paral{\semProb_j}_{j\neq\text{background}},
\end{equation}
where $\semProb_j \in \zeroOne^C$ provided the scores for $C = 8$ findings per tooth index $j$.
For 32 tooth indices, the system produced a total of 256 floating-point scores per DPR.

\newpage

\section{AI Performance -- Detailed Results}
\label{sec:performance-comparison-detailed}
\input{./includes/tbl-ai-multinational}

\newpage

\section{Consistency of Human Reader Performance}
\label{sec:consistency-of-human}
\input{./includes/fig-reader-agreement}

We inspected the mutual agreement among four readers (G1, G2, S1, and S2) involved in the study using Cohen's Kappa.
Kappa was calculated between every possible pair of human readers, and the overall average of agreements was strong~\citep{mchugh2012interrater} for the following findings: crown/bridges at 95.4\% (95\% CI: 92.9\% -- 97.8\%), implants at 92.3\% (95\% CI: 83.0\% -- 100.0\%), root canal fillings at 92.1\% (95\% CI: 86.8\% -- 97.4\%), and missing teeth at 90.1\% (95\% CI: 81.9\% -- 98.2\%).
Agreement for fillings was moderate, at 74.9\% (95\% CI: 69.8\% -- 80.0\%).
Residual roots and caries exhibited weak agreement, at 56.2\% (95\% CI: 35.6\% -- 76.8\%) and 46.5\% (95\% CI: 34.8\% -- 58.1\%), respectively.
The lowest mean agreement occurred for periapical radiolucencies, at 37.6\% (95\% CI: 14.7\% -- 60.6\%).

To further analyze the data, we categorized the readers into two expertise groups: general dentists (G1 and G2) and specialized dentists (S1 and S2).
Average agreement metrics were computed by taking the average Kappa value over reader pairs with specific criteria: \emph{same-expertise} pairs (G1-G2 and S1-S2) and \emph{different-expertise} pairs (G1-S1, G1-S2, G2-S1, and G2-S2), respectively.
Notably, for residual roots, a statistically significant difference in agreement was observed between the same-expertise and different-expertise average Kappa values, with a difference of 30.3\% (95\% CI: 1.6\% -- 58.9\%, $p = .039$).
No significant differences were found in agreement levels between same-expertise average and different-expertise average for other findings.

%% file: includes/tbl-ai-multinational.tex
\begin{table}[!h]
\centering\caption{  \textbf{Comprehensive Performance Comparison of AI and Human Readers.}   This table presents a horizontal comparison of various performance metrics for the AI system and human readers across multiple test datasets and prior studies.  The leftmost columns show results for the AI system on different test sets in this study, where the AI was optimized to maximize the $\textrm{F}_2$ score in a screening setting.  Note that Taiwan* represents a subset of the Taiwan test set.  The rightmost columns summarize performance metrics reported in prior works, evaluated using their respective datasets and AI systems.  Values in parentheses indicate the 95\% confidence intervals.  A ``--'' indicates that a finding was not included, the metric was not reported, or the metric was not applicable.}\label{tbl:ai-multinational}\tiny\rowcolors{1}{white}{table-bg}
\begin{tabular}{ p{0.167\linewidth}p{0.117\linewidth}p{0.083\linewidth}p{0.083\linewidth}p{0.083\linewidth}p{0.083\linewidth}|p{0.067\linewidth}p{0.067\linewidth} }
\toprule
Data Set & The Netherlands, test & Brazil & Taiwan & Taiwan* & Taiwan* & \citet{vinayahalingam2021automated} & \citet{bacsaran2022diagnostic} \\ \midrule
Reader & AI (Ours) & AI (Ours) & AI (Ours) & AI (Ours) & Human, mean & AI~\citep{vinayahalingam2021automated} & AI~\citep{bacsaran2022diagnostic} \\ \midrule
Missing &   &   &   &   &   &   &   \\
{~~~~}Positive/negative count & \num{6836} / \num{31596} & \num{5926} / \num{31610} & \num{1819} / \num{6213} & \num{866} / \num{2910} & \num{866} / \num{2910} & -- & -- \\
{~~~~}Metric, \% (95\% CI) &   &   &   &   &   &   &   \\
{~~~~~~~~}Sensitivity & 87.6 {(86.8--88.4)} & 93.6 {(92.9--94.2)} & 89.3 {(87.8--90.7)} & 92.0 {(90.0--93.7)} & 87.3 {(85.5--89.1)} & 96.5 & -- \\
{~~~~~~~~}Specificity & 95.5 {(95.3--95.8)} & 95.7 {(95.5--95.9)} & 95.1 {(94.5--95.6)} & 95.9 {(95.1--96.5)} & 99.1 {(98.9--99.4)} & -- & -- \\
{~~~~~~~~}Precision & 80.9 {(80.0--81.8)} & 80.3 {(79.3--81.2)} & 84.2 {(82.5--85.7)} & 86.9 {(84.6--88.9)} & 96.8 {(95.9--97.7)} & 97.5 & -- \\
{~~~~~~~~}$\textrm{F}_1$ score & 84.2 {(83.5--84.9)} & 86.4 {(85.7--87.1)} & 86.7 {(85.4--87.9)} & 89.4 {(87.7--91.1)} & 91.8 {(90.7--92.9)} & 97.0 & -- \\
{~~~~~~~~}$\textrm{F}_2$ score & 86.2 {(85.5--86.9)} & 90.6 {(90.0--91.2)} & 88.2 {(86.9--89.6)} & 91.0 {(89.3--92.7)} & 89.1 {(87.6--90.6)} & -- & -- \\
{~~~~~~~~}AUC-ROC & 94.4 {(94.0--94.9)} & 97.8 {(97.6--98.1)} & 95.7 {(95.1--96.4)} & 96.9 {(96.1--97.7)} & -- & -- & -- \\ \midrule

Implant &   &   &   &   &   &   &   \\
{~~~~}Positive/negative count & \num{619} / \num{6217} & \num{586} / \num{5340} & \num{191} / \num{1628} & \num{80} / \num{786} & \num{80} / \num{786} & -- & \num{26} / -- \\
{~~~~}Metric, \% (95\% CI) &   &   &   &   &   &   &   \\
{~~~~~~~~}Sensitivity & 98.2 {(96.8--99.0)} & 95.6 {(93.6--97.0)} & 99.0 {(96.3--99.7)} & 100.0 {(95.4--100.0)} & 95.0 {(92.3--97.7)} & 81.7 & 96.15 \\
{~~~~~~~~}Specificity & 97.6 {(97.2--98.0)} & 95.0 {(94.4--95.6)} & 98.3 {(97.5--98.8)} & 98.3 {(97.2--99.0)} & 99.7 {(99.4--99.9)} & -- & -- \\
{~~~~~~~~}Precision & 80.3 {(77.3--83.0)} & 67.8 {(64.5--70.9)} & 87.1 {(82.0--90.9)} & 86.0 {(77.5--91.6)} & 96.8 {(94.3--99.3)} & 80.0 & 92.59 \\
{~~~~~~~~}$\textrm{F}_1$ score & 88.4 {(86.5--90.2)} & 79.3 {(76.9--81.7)} & 92.6 {(89.8--95.5)} & 92.5 {(88.2--96.8)} & 95.9 {(94.0--97.8)} & 80.9 & 94.33 \\
{~~~~~~~~}$\textrm{F}_2$ score & 94.0 {(92.8--95.2)} & 88.3 {(86.6--90.1)} & 96.3 {(94.6--98.1)} & 96.9 {(94.8--98.9)} & 95.3 {(93.1--97.6)} & -- & -- \\
{~~~~~~~~}AUC-ROC & 99.5 {(99.4--99.6)} & 98.3 {(98.1--98.6)} & 99.7 {(99.6--99.9)} & 99.8 {(99.6--99.9)} & -- & -- & -- \\ \midrule
 
Residual root &   &   &   &   &   &   &   \\
{~~~~}Positive/negative count & \num{137} / \num{31459} & \num{198} / \num{31412} & \num{88} / \num{6125} & \num{38} / \num{2872} & \num{38} / \num{2872} & -- & \num{29} / -- \\
{~~~~}Metric, \% (95\% CI) &   &   &   &   &   &   &   \\
{~~~~~~~~}Sensitivity & 79.6 {(72.0--85.5)} & 56.1 {(49.1--62.8)} & 72.7 {(62.6--80.9)} & 71.1 {(55.2--83.0)} & 57.2 {(48.9--65.6)} & 73.3 & 82.14 \\
{~~~~~~~~}Specificity & 99.9 {(99.8--99.9)} & 99.5 {(99.4--99.6)} & 99.0 {(98.8--99.3)} & 99.2 {(98.8--99.4)} & 99.9 {(99.8--100.0)} & -- & -- \\
{~~~~~~~~}Precision & 76.2 {(68.6--82.5)} & 41.7 {(36.0--47.7)} & 52.0 {(43.3--60.7)} & 52.9 {(39.5--65.9)} & 90.0 {(79.7--100.4)} & 81.5 & 67.64 \\
{~~~~~~~~}$\textrm{F}_1$ score & 77.9 {(72.5--83.2)} & 47.8 {(42.2--53.5)} & 60.7 {(52.9--68.4)} & 60.7 {(48.7--72.7)} & 67.8 {(59.4--76.1)} & 77.2 & 74.19 \\
{~~~~~~~~}$\textrm{F}_2$ score & 78.9 {(73.0--84.7)} & 52.5 {(46.4--58.5)} & 67.4 {(59.4--75.3)} & 66.5 {(54.1--78.9)} & 60.8 {(52.5--69.1)} & -- & -- \\
{~~~~~~~~}AUC-ROC & 93.1 {(89.8--96.4)} & 92.8 {(90.3--95.3)} & 98.3 {(97.4--99.2)} & 97.6 {(95.7--99.5)} & -- & -- & -- \\ \midrule

Crown/bridge &   &   &   &   &   &   &   \\
{~~~~}Positive/negative count & \num{3038} / \num{28558} & \num{896} / \num{30714} & \num{851} / \num{5362} & \num{429} / \num{2481} & \num{429} / \num{2481} & -- & \num{236} / -- \\
{~~~~}Metric, \% (95\% CI) &   &   &   &   &   &   &   \\
{~~~~~~~~}Sensitivity & 92.7 {(91.7--93.5)} & 87.4 {(85.1--89.4)} & 97.2 {(95.8--98.1)} & 98.1 {(96.4--99.1)} & 96.4 {(95.2--97.6)} & 85.1 & 96.74 \\
{~~~~~~~~}Specificity & 98.7 {(98.6--98.8)} & 98.8 {(98.7--98.9)} & 99.4 {(99.1--99.6)} & 98.9 {(98.4--99.2)} & 99.7 {(99.6--99.8)} & -- & -- \\
{~~~~~~~~}Precision & 88.5 {(87.4--89.6)} & 68.3 {(65.5--70.9)} & 96.2 {(94.7--97.3)} & 93.8 {(91.1--95.7)} & 98.2 {(97.4--99.0)} & 87.2 & 86.30 \\
{~~~~~~~~}$\textrm{F}_1$ score & 90.6 {(89.8--91.4)} & 76.7 {(74.6--78.7)} & 96.7 {(95.7--97.6)} & 95.9 {(94.4--97.4)} & 97.3 {(96.5--98.0)} & 86.1 & 91.22 \\
{~~~~~~~~}$\textrm{F}_2$ score & 91.8 {(91.0--92.6)} & 82.8 {(80.8--84.7)} & 97.0 {(96.0--98.0)} & 97.2 {(96.0--98.5)} & 96.7 {(95.8--97.7)} & -- & -- \\
{~~~~~~~~}AUC-ROC & 98.7 {(98.5--98.9)} & 98.0 {(97.4--98.6)} & 99.8 {(99.7--99.9)} & 99.8 {(99.7--99.9)} & -- & -- & -- \\ \midrule

Root canal filling &   &   &   &   &   &   &   \\
{~~~~}Positive/negative count & \num{1867} / \num{29729} & \num{1316} / \num{30294} & \num{643} / \num{5570} & \num{326} / \num{2584} & \num{326} / \num{2584} & -- & \num{162} / -- \\
{~~~~}Metric, \% (95\% CI) &   &   &   &   &   &   &   \\
{~~~~~~~~}Sensitivity & 97.4 {(96.5--98.0)} & 94.1 {(92.7--95.3)} & 96.7 {(95.1--97.9)} & 97.2 {(94.8--98.5)} & 93.3 {(91.6--95.0)} & 88.2 & 86.70 \\
{~~~~~~~~}Specificity & 99.6 {(99.5--99.7)} & 99.4 {(99.3--99.5)} & 99.3 {(99.0--99.5)} & 99.3 {(98.9--99.6)} & 99.8 {(99.7--99.9)} & -- & -- \\
{~~~~~~~~}Precision & 93.6 {(92.4--94.6)} & 86.8 {(84.9--88.4)} & 93.8 {(91.7--95.4)} & 94.6 {(91.7--96.6)} & 98.1 {(97.2--99.0)} & 89.1 & 77.72 \\
{~~~~~~~~}$\textrm{F}_1$ score & 95.5 {(94.8--96.2)} & 90.3 {(89.1--91.5)} & 95.3 {(94.0--96.5)} & 95.9 {(94.2--97.6)} & 95.6 {(94.6--96.6)} & 88.6 & 81.96 \\
{~~~~~~~~}$\textrm{F}_2$ score & 96.6 {(95.9--97.3)} & 92.6 {(91.5--93.7)} & 96.1 {(94.9--97.4)} & 96.7 {(95.1--98.3)} & 94.2 {(92.8--95.6)} & -- & -- \\
{~~~~~~~~}AUC-ROC & 99.5 {(99.3--99.7)} & 99.2 {(99.0--99.5)} & 99.5 {(99.1--99.8)} & 99.4 {(98.8--100.0)} & -- & -- & -- \\ \midrule

Filling &   &   &   &   &   &   &   \\
{~~~~}Positive/negative count & \num{7412} / \num{24184} & \num{8127} / \num{23483} & \num{1370} / \num{4843} & \num{611} / \num{2299} & \num{611} / \num{2299} & -- & \num{519} / -- \\
{~~~~}Metric, \% (95\% CI) &   &   &   &   &   &   &   \\
{~~~~~~~~}Sensitivity & 92.2 {(91.6--92.8)} & 93.8 {(93.3--94.3)} & 87.6 {(85.7--89.2)} & 84.8 {(81.7--87.4)} & 74.4 {(71.8--77.0)} & 81.9 & 86.08 \\
{~~~~~~~~}Specificity & 93.9 {(93.6--94.2)} & 92.3 {(91.9--92.6)} & 88.0 {(87.1--88.9)} & 94.6 {(93.6--95.5)} & 98.4 {(98.1--98.7)} & -- & -- \\
{~~~~~~~~}Precision & 82.3 {(81.5--83.1)} & 80.8 {(80.0--81.6)} & 67.4 {(65.2--69.6)} & 80.7 {(77.5--83.6)} & 93.1 {(91.8--94.4)} & 84.1 & 87.61 \\
{~~~~~~~~}$\textrm{F}_1$ score & 87.0 {(86.4--87.6)} & 86.8 {(86.2--87.4)} & 76.2 {(74.5--77.9)} & 82.7 {(80.2--85.1)} & 82.1 {(80.3--83.9)} & 83.0 & 86.84 \\
{~~~~~~~~}$\textrm{F}_2$ score & 90.0 {(89.5--90.6)} & 90.9 {(90.4--91.4)} & 82.6 {(81.0--84.3)} & 83.9 {(81.3--86.5)} & 77.2 {(74.9--79.5)} & -- & -- \\
{~~~~~~~~}AUC-ROC & 97.3 {(97.0--97.5)} & 97.2 {(97.0--97.4)} & 94.8 {(94.1--95.5)} & 95.9 {(95.0--96.8)} & -- & -- & -- \\ \midrule

Caries &   &   &   &   &   &   &   \\
{~~~~}Positive/negative count & \num{1063} / \num{30533} & \num{1435} / \num{30175} & \num{317} / \num{5896} & \num{180} / \num{2730} & \num{180} / \num{2730} & -- & \num{256} / -- \\
{~~~~}Metric, \% (95\% CI) &   &   &   &   &   &   &   \\
{~~~~~~~~}Sensitivity & 46.3 {(43.3--49.3)} & 57.4 {(54.8--59.9)} & 52.7 {(47.2--58.1)} & 56.1 {(48.8--63.2)} & 40.6 {(35.3--45.8)} & -- & 30.26 \\
{~~~~~~~~}Specificity & 91.8 {(91.5--92.1)} & 84.7 {(84.3--85.1)} & 94.5 {(93.9--95.1)} & 95.3 {(94.5--96.0)} & 99.0 {(98.8--99.2)} & -- & -- \\
{~~~~~~~~}Precision & 16.5 {(15.2--17.9)} & 15.2 {(14.2--16.1)} & 34.1 {(30.0--38.4)} & 44.1 {(37.8--50.6)} & 77.4 {(72.3--82.4)} & -- & 50.96 \\
{~~~~~~~~}$\textrm{F}_1$ score & 24.3 {(22.5--26.1)} & 24.0 {(22.6--25.4)} & 41.4 {(37.1--45.7)} & 49.4 {(43.4--55.3)} & 50.7 {(45.3--56.1)} & -- & 37.98 \\
{~~~~~~~~}$\textrm{F}_2$ score & 34.0 {(31.8--36.2)} & 36.8 {(35.1--38.6)} & 47.5 {(42.8--52.2)} & 53.2 {(46.9--59.6)} & 43.9 {(38.6--49.2)} & -- & -- \\
{~~~~~~~~}AUC-ROC & 82.5 {(81.3--83.7)} & 79.8 {(78.6--81.0)} & 82.3 {(79.6--84.9)} & 83.0 {(79.3--86.6)} & -- & -- & -- \\ \midrule

Periapical radiolucency &   &   &   &   &   &   &   \\
{~~~~}Positive/negative count & \num{458} / \num{31138} & \num{377} / \num{31233} & \num{197} / \num{6016} & \num{138} / \num{2772} & \num{138} / \num{2772} & -- & -- \\
{~~~~}Metric, \% (95\% CI) &   &   &   &   &   &   &   \\
{~~~~~~~~}Sensitivity & 55.2 {(50.7--59.7)} & 53.8 {(48.8--58.8)} & 86.8 {(81.4--90.8)} & 93.5 {(88.1--96.5)} & 25.5 {(20.6--30.5)} & -- & -- \\
{~~~~~~~~}Specificity & 98.2 {(98.1--98.3)} & 97.2 {(97.1--97.4)} & 94.9 {(94.3--95.4)} & 97.0 {(96.3--97.6)} & 99.5 {(99.4--99.7)} & -- & -- \\
{~~~~~~~~}Precision & 31.2 {(28.1--34.4)} & 19.1 {(16.9--21.6)} & 35.8 {(31.6--40.2)} & 61.1 {(54.4--67.5)} & 73.1 {(64.1--82.1)} & -- & -- \\
{~~~~~~~~}$\textrm{F}_1$ score & 39.8 {(36.4--43.3)} & 28.2 {(25.2--31.3)} & 50.7 {(46.1--55.3)} & 73.9 {(68.7--79.1)} & 37.0 {(30.9--43.0)} & -- & -- \\
{~~~~~~~~}$\textrm{F}_2$ score & 47.8 {(44.1--51.6)} & 39.5 {(35.7--43.3)} & 67.5 {(63.2--71.8)} & 84.5 {(80.5--88.6)} & 29.1 {(23.7--34.5)} & -- & -- \\
{~~~~~~~~}AUC-ROC & 93.2 {(92.0--94.5)} & 88.1 {(85.9--90.3)} & 95.0 {(93.3--96.8)} & 97.4 {(95.9--99.0)} & -- & -- & -- \\ \bottomrule
\end{tabular}
\end{table}

%% file: includes/fig-reader-agreement.tex
\begin{figure}[!t]
    \centering
    \includegraphics[width=1.0\textwidth, height=0.80\textheight, keepaspectratio]{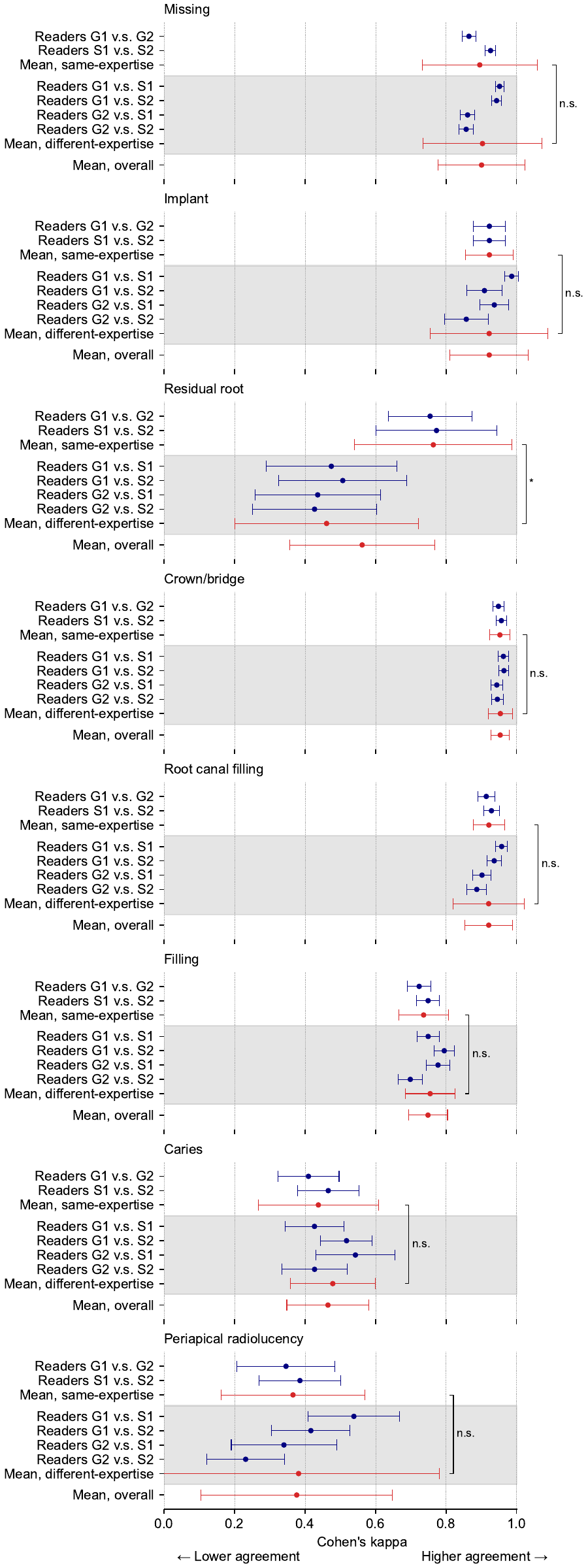}
    \caption{
        \textbf{Inter-Reader Agreement Levels for Dental Finding Summaries in Taiwan.}
        This figure presents the agreement levels between pairs of readers as measured by Cohen's Kappa for various dental findings.
        Each blue error bar illustrates the Kappa agreement between each of the six possible reader pairings (G1/G2, G1/S1, G1/S2, G2/S1, G2/S2, and S1/S2) from four participating readers, grouped into general dentists (G1 and G2) and specialists (S1 and S2).
        The agreement levels are averaged for pairs within the same expertise group (generalists or specialists) and across different expertise, shown as red lines.
        The error bars represent the 95\% confidence intervals for the Kappa values.
        Significance testing using $t$-statistics revealed that differences in mean agreement levels for residual roots were statistically significant ($p = .039$), while other findings showed no significant differences.
    }
    \label{fig:reader-agreement}
\end{figure}

%% file: main.bbl
\begin{thebibliography}{37}
\providecommand{\natexlab}[1]{#1}
\providecommand{\url}[1]{\texttt{#1}}
\expandafter\ifx\csname urlstyle\endcsname\relax
  \providecommand{\doi}[1]{doi: #1}\else
  \providecommand{\doi}{doi: \begingroup \urlstyle{rm}\Url}\fi

\bibitem[Choi(2011)]{choi2011assessment}
Jin-Woo Choi.
\newblock Assessment of panoramic radiography as a national oral examination tool: review of the literature.
\newblock \emph{Imaging science in dentistry}, 41\penalty0 (1):\penalty0 1, 2011.

\bibitem[Fourcade and Khonsari(2019)]{fourcade2019deep}
Arthur Fourcade and Roman~Hossein Khonsari.
\newblock Deep learning in medical image analysis: A third eye for doctors.
\newblock \emph{Journal of stomatology, oral and maxillofacial surgery}, 120\penalty0 (4):\penalty0 279--288, 2019.

\bibitem[Kweon et~al.(2018)Kweon, Lee, Youk, Lee, and Kim]{kweon2018panoramic}
Helen Hye-In Kweon, Jae-Hong Lee, Tae-mi Youk, Bo-Ah Lee, and Young-Taek Kim.
\newblock Panoramic radiography can be an effective diagnostic tool adjunctive to oral examinations in the national health checkup program.
\newblock \emph{Journal of periodontal \& implant science}, 48\penalty0 (5):\penalty0 317, 2018.

\bibitem[Plessas et~al.(2019)Plessas, Nasser, Hanoch, O'Brien, Delgado, and Moles]{plessas2019impact}
Anastasios Plessas, Mona Nasser, Yaniv Hanoch, Timothy O'Brien, Maria~Bernardes Delgado, and David Moles.
\newblock Impact of time pressure on dentists' diagnostic performance.
\newblock \emph{Journal of dentistry}, 82:\penalty0 38--44, 2019.

\bibitem[Oktay(2017)]{oktay2017tooth}
Ayse~Betul Oktay.
\newblock Tooth detection with convolutional neural networks.
\newblock In \emph{2017 Medical Technologies National Congress (TIPTEKNO)}, pages 1--4. IEEE, 2017.

\bibitem[Koch et~al.(2019)Koch, Perslev, Igel, and Brandt]{koch2019accurate}
Thorbj{\o}rn~Louring Koch, Mathis Perslev, Christian Igel, and Sami~Sebastian Brandt.
\newblock Accurate segmentation of dental panoramic radiographs with u-nets.
\newblock In \emph{2019 IEEE 16th International Symposium on Biomedical Imaging (ISBI 2019)}, pages 15--19. IEEE, 2019.

\bibitem[Tuzoff et~al.(2019)Tuzoff, Tuzova, Bornstein, Krasnov, Kharchenko, Nikolenko, Sveshnikov, and Bednenko]{tuzoff2019tooth}
Dmitry~V Tuzoff, Lyudmila~N Tuzova, Michael~M Bornstein, Alexey~S Krasnov, Max~A Kharchenko, Sergey~I Nikolenko, Mikhail~M Sveshnikov, and Georgiy~B Bednenko.
\newblock Tooth detection and numbering in panoramic radiographs using convolutional neural networks.
\newblock \emph{Dentomaxillofacial Radiology}, 48\penalty0 (4):\penalty0 20180051, 2019.

\bibitem[Kim et~al.(2020)Kim, Kim, Jeong, Yoon, and Youm]{kim2020automatic}
Changgyun Kim, Donghyun Kim, HoGul Jeong, Suk-Ja Yoon, and Sekyoung Youm.
\newblock Automatic tooth detection and numbering using a combination of a cnn and heuristic algorithm.
\newblock \emph{Applied Sciences}, 10\penalty0 (16):\penalty0 5624, 2020.

\bibitem[Xu et~al.(2023)Xu, Wu, Xu, Ding, Bai, and Deng]{xu2023robust}
Mingming Xu, Yujia Wu, Zineng Xu, Peng Ding, Hailong Bai, and Xuliang Deng.
\newblock Robust automated teeth identification from dental radiographs using deep learning.
\newblock \emph{Journal of Dentistry}, 136:\penalty0 104607, 2023.

\bibitem[Day{\i} et~al.(2023)Day{\i}, {\"U}zen, {\c{C}}i{\c{c}}ek, and Duman]{dayi2023novel}
Burak Day{\i}, H{\"u}seyin {\"U}zen, {\.I}pek~Bal{\i}k{\c{c}}{\i} {\c{C}}i{\c{c}}ek, and {\c{S}}uayip~Burak Duman.
\newblock A novel deep learning-based approach for segmentation of different type caries lesions on panoramic radiographs.
\newblock \emph{Diagnostics}, 13\penalty0 (2):\penalty0 202, 2023.

\bibitem[Zhu et~al.(2022)Zhu, Cao, Lian, Ye, Gao, and Wu]{zhu2022cariesnet}
Haihua Zhu, Zheng Cao, Luya Lian, Guanchen Ye, Honghao Gao, and Jian Wu.
\newblock Cariesnet: a deep learning approach for segmentation of multi-stage caries lesion from oral panoramic x-ray image.
\newblock \emph{Neural Computing and Applications}, pages 1--9, 2022.

\bibitem[Oztekin et~al.(2023)Oztekin, Katar, Sadak, Yildirim, Cakar, Aydogan, Ozpolat, Talo~Yildirim, Yildirim, Faust, et~al.]{oztekin2023explainable}
Faruk Oztekin, Oguzhan Katar, Ferhat Sadak, Muhammed Yildirim, Hakan Cakar, Murat Aydogan, Zeynep Ozpolat, Tuba Talo~Yildirim, Ozal Yildirim, Oliver Faust, et~al.
\newblock An explainable deep learning model to prediction dental caries using panoramic radiograph images.
\newblock \emph{Diagnostics}, 13\penalty0 (2):\penalty0 226, 2023.

\bibitem[Lian et~al.(2021)Lian, Zhu, Zhu, and Zhu]{lian2021deep}
Luya Lian, Tianer Zhu, Fudong Zhu, and Haihua Zhu.
\newblock Deep learning for caries detection and classification.
\newblock \emph{Diagnostics}, 11\penalty0 (9):\penalty0 1672, 2021.

\bibitem[Endres et~al.(2020)Endres, Hillen, Salloumis, Sedaghat, Niehues, Quatela, Hanken, Smeets, Beck-Broichsitter, Rendenbach, et~al.]{endres2020development}
Michael~G Endres, Florian Hillen, Marios Salloumis, Ahmad~R Sedaghat, Stefan~M Niehues, Olivia Quatela, Henning Hanken, Ralf Smeets, Benedicta Beck-Broichsitter, Carsten Rendenbach, et~al.
\newblock Development of a deep learning algorithm for periapical disease detection in dental radiographs.
\newblock \emph{Diagnostics}, 10\penalty0 (6):\penalty0 430, 2020.

\bibitem[Yang et~al.(2020)Yang, Jo, Kim, Cha, Jung, Nam, Kim, Kim, Kim, Oh, et~al.]{yang2020deep}
Hyunwoo Yang, Eun Jo, Hyung~Jun Kim, In-ho Cha, Young-Soo Jung, Woong Nam, Jun-Young Kim, Jin-Kyu Kim, Yoon~Hyeon Kim, Tae~Gyeong Oh, et~al.
\newblock Deep learning for automated detection of cyst and tumors of the jaw in panoramic radiographs.
\newblock \emph{Journal of clinical medicine}, 9\penalty0 (6):\penalty0 1839, 2020.

\bibitem[Muresan et~al.(2020)Muresan, Barbura, and Nedevschi]{muresan2020teeth}
Mircea~Paul Muresan, Andrei~R{\u{a}}zvan Barbura, and Sergiu Nedevschi.
\newblock Teeth detection and dental problem classification in panoramic x-ray images using deep learning and image processing techniques.
\newblock In \emph{2020 IEEE 16th International Conference on Intelligent Computer Communication and Processing (ICCP)}, pages 457--463. IEEE, 2020.

\bibitem[Vinayahalingam et~al.(2021)Vinayahalingam, Goey, Kempers, Schoep, Cherici, Moin, and Hanisch]{vinayahalingam2021automated}
Shankeeth Vinayahalingam, Ru-shan Goey, Steven Kempers, Julian Schoep, Teo Cherici, David~Anssari Moin, and Marcel Hanisch.
\newblock Automated chart filing on panoramic radiographs using deep learning.
\newblock \emph{Journal of Dentistry}, 115:\penalty0 103864, 2021.

\bibitem[Rohrer et~al.(2022)Rohrer, Krois, Patel, Meyer-Lueckel, Rodrigues, and Schwendicke]{rohrer2022segmentation}
Csaba Rohrer, Joachim Krois, Jay Patel, Hendrik Meyer-Lueckel, Jonas~Almeida Rodrigues, and Falk Schwendicke.
\newblock Segmentation of dental restorations on panoramic radiographs using deep learning.
\newblock \emph{Diagnostics}, 12\penalty0 (6):\penalty0 1316, 2022.

\bibitem[Almalki et~al.(2022)Almalki, Din, Ramzan, Irfan, Aamir, Almalki, Alotaibi, Alaglan, Alshamrani, and Rahman]{almalki2022deep}
Yassir~Edrees Almalki, Amsa~Imam Din, Muhammad Ramzan, Muhammad Irfan, Khalid~Mahmood Aamir, Abdullah Almalki, Saud Alotaibi, Ghada Alaglan, Hassan~A Alshamrani, and Saifur Rahman.
\newblock Deep learning models for classification of dental diseases using orthopantomography x-ray opg images.
\newblock \emph{Sensors}, 22\penalty0 (19):\penalty0 7370, 2022.

\bibitem[Ba{\c{s}}aran et~al.(2022)Ba{\c{s}}aran, {\c{C}}elik, Bayrakdar, Bilgir, Orhan, Odaba{\c{s}}, Aslan, and Jagtap]{bacsaran2022diagnostic}
Melike Ba{\c{s}}aran, {\"O}zer {\c{C}}elik, Ibrahim~Sevki Bayrakdar, Elif Bilgir, Kaan Orhan, Alper Odaba{\c{s}}, Ahmet~Faruk Aslan, and Rohan Jagtap.
\newblock Diagnostic charting of panoramic radiography using deep-learning artificial intelligence system.
\newblock \emph{Oral radiology}, pages 1--7, 2022.

\bibitem[Gardiyano{\u{g}}lu et~al.(2023)Gardiyano{\u{g}}lu, {\"U}nsal, Akkaya, Aksoy, and Orhan]{gardiyanouglu2023automatic}
Emel Gardiyano{\u{g}}lu, G{\"u}rkan {\"U}nsal, Nurullah Akkaya, Se{\c{c}}il Aksoy, and Kaan Orhan.
\newblock Automatic segmentation of teeth, crown--bridge restorations, dental implants, restorative fillings, dental caries, residual roots, and root canal fillings on orthopantomographs: Convenience and pitfalls.
\newblock \emph{Diagnostics}, 13\penalty0 (8):\penalty0 1487, 2023.

\bibitem[van Nistelrooij et~al.(2024)van Nistelrooij, Ghoul, Xi, Saha, Kempers, Cenci, Loomans, Fl{\"u}gge, van Ginneken, and Vinayahalingam]{van2024combining}
Niels van Nistelrooij, Khalid~El Ghoul, Tong Xi, Anindo Saha, Steven Kempers, Max Cenci, Bas Loomans, Tabea Fl{\"u}gge, Bram van Ginneken, and Shankeeth Vinayahalingam.
\newblock Combining public datasets for automated tooth assessment in panoramic radiographs.
\newblock \emph{BMC Oral Health}, 24\penalty0 (1):\penalty0 387, 2024.

\bibitem[Silva et~al.(2018)Silva, Oliveira, and Pithon]{silva2018automatic}
Gil Silva, Luciano Oliveira, and Matheus Pithon.
\newblock Automatic segmenting teeth in {X}-ray images: Trends, a novel data set, benchmarking and future perspectives.
\newblock \emph{Expert Systems with Applications}, 107:\penalty0 15--31, 2018.

\bibitem[Redmon et~al.(2016)Redmon, Divvala, Girshick, and Farhadi]{redmon2016you}
Joseph Redmon, Santosh Divvala, Ross Girshick, and Ali Farhadi.
\newblock You only look once: Unified, real-time object detection.
\newblock In \emph{Proceedings of the IEEE conference on computer vision and pattern recognition}, pages 779--788, 2016.

\bibitem[Chen et~al.(2017)Chen, Papandreou, Kokkinos, Murphy, and Yuille]{chen2017deeplab}
Liang-Chieh Chen, George Papandreou, Iasonas Kokkinos, Kevin Murphy, and Alan~L Yuille.
\newblock Deeplab: Semantic image segmentation with deep convolutional nets, atrous convolution, and fully connected crfs.
\newblock \emph{IEEE transactions on pattern analysis and machine intelligence}, 40\penalty0 (4):\penalty0 834--848, 2017.

\bibitem[DeLong et~al.(1988)DeLong, DeLong, and Clarke-Pearson]{delong1988comparing}
Elizabeth~R DeLong, David~M DeLong, and Daniel~L Clarke-Pearson.
\newblock Comparing the areas under two or more correlated receiver operating characteristic curves: a nonparametric approach.
\newblock \emph{Biometrics}, pages 837--845, 1988.

\bibitem[McHugh(2012)]{mchugh2012interrater}
Mary~L McHugh.
\newblock Interrater reliability: the kappa statistic.
\newblock \emph{Biochemia medica}, 22\penalty0 (3):\penalty0 276--282, 2012.

\bibitem[McKinney(2022)]{mckinney2022comparing}
Scott~Mayer McKinney.
\newblock Comparing human and ai performance in medical machine learning: An open-source python library for the statistical analysis of reader study data.
\newblock \emph{medRxiv}, pages 2022--05, 2022.

\bibitem[Muramatsu et~al.(2021)Muramatsu, Morishita, Takahashi, Hayashi, Nishiyama, Ariji, Zhou, Hara, Katsumata, Ariji, et~al.]{muramatsu2021tooth}
Chisako Muramatsu, Takumi Morishita, Ryo Takahashi, Tatsuro Hayashi, Wataru Nishiyama, Yoshiko Ariji, Xiangrong Zhou, Takeshi Hara, Akitoshi Katsumata, Eiichiro Ariji, et~al.
\newblock Tooth detection and classification on panoramic radiographs for automatic dental chart filing: improved classification by multi-sized input data.
\newblock \emph{Oral Radiology}, 37:\penalty0 13--19, 2021.

\bibitem[Leite et~al.(2021)Leite, Gerven, Willems, Beznik, Lahoud, Ga{\^e}ta-Araujo, Vranckx, and Jacobs]{leite2021artificial}
Andr{\'e}~Ferreira Leite, Adriaan~Van Gerven, Holger Willems, Thomas Beznik, Pierre Lahoud, Hugo Ga{\^e}ta-Araujo, Myrthel Vranckx, and Reinhilde Jacobs.
\newblock Artificial intelligence-driven novel tool for tooth detection and segmentation on panoramic radiographs.
\newblock \emph{Clinical oral investigations}, 25:\penalty0 2257--2267, 2021.

\bibitem[Chen et~al.(2022)Chen, Chen, Huang, Chen, Chou, Huang, Lin, Li, Yuan, Abu, et~al.]{chen2022missing}
Shih-Lun Chen, Tsung-Yi Chen, Yen-Cheng Huang, Chiung-An Chen, He-Sheng Chou, Ya-Yun Huang, Wei-Chi Lin, Tzu-Chien Li, Jia-Jun Yuan, Patricia Angela~R Abu, et~al.
\newblock Missing teeth and restoration detection using dental panoramic radiography based on transfer learning with cnns.
\newblock \emph{IEEE Access}, 10:\penalty0 118654--118664, 2022.

\bibitem[Liang et~al.(2022)Liang, Tsai, Shih, Huang, Morisky, and Fu]{liang2022tooth}
Ling-Yu Liang, Ming-Che Tsai, Kuang-Chung Shih, Shiu-Ming Huang, Donald~E Morisky, and Earl Fu.
\newblock Tooth life expectancy and burden of tooth loss: Two cross-sectional studies in taiwan.
\newblock \emph{Journal of Dental Sciences}, 17\penalty0 (3):\penalty0 1364--1370, 2022.

\bibitem[Varghese and Sambath(2024)]{varghese2024yolov8}
Rejin Varghese and M~Sambath.
\newblock Yolov8: A novel object detection algorithm with enhanced performance and robustness.
\newblock In \emph{2024 International Conference on Advances in Data Engineering and Intelligent Computing Systems (ADICS)}, pages 1--6. IEEE, 2024.

\bibitem[Lin et~al.(2014)Lin, Maire, Belongie, Hays, Perona, Ramanan, Doll{\'a}r, and Zitnick]{lin2014microsoft}
Tsung-Yi Lin, Michael Maire, Serge Belongie, James Hays, Pietro Perona, Deva Ramanan, Piotr Doll{\'a}r, and C~Lawrence Zitnick.
\newblock Microsoft coco: Common objects in context.
\newblock In \emph{Computer Vision--ECCV 2014: 13th European Conference, Zurich, Switzerland, September 6-12, 2014, Proceedings, Part V 13}, pages 740--755. Springer, 2014.

\bibitem[Chen(2017)]{chen2017rethinking}
Liang-Chieh Chen.
\newblock Rethinking atrous convolution for semantic image segmentation.
\newblock \emph{arXiv preprint arXiv:1706.05587}, 2017.

\bibitem[Wu et~al.(2019)Wu, Kirillov, Massa, Lo, and Girshick]{wu2019detectron2}
Yuxin Wu, Alexander Kirillov, Francisco Massa, Wan-Yen Lo, and Ross Girshick.
\newblock Detectron2.
\newblock \url{https://github.com/facebookresearch/detectron2}, 2019.

\bibitem[Ardila et~al.(2019)Ardila, Kiraly, Bharadwaj, Choi, Reicher, Peng, Tse, Etemadi, Ye, Corrado, et~al.]{ardila2019end}
Diego Ardila, Atilla~P Kiraly, Sujeeth Bharadwaj, Bokyung Choi, Joshua~J Reicher, Lily Peng, Daniel Tse, Mozziyar Etemadi, Wenxing Ye, Greg Corrado, et~al.
\newblock End-to-end lung cancer screening with three-dimensional deep learning on low-dose chest computed tomography.
\newblock \emph{Nature medicine}, 25\penalty0 (6):\penalty0 954--961, 2019.

\end{thebibliography}
